\documentclass{article}

\PassOptionsToPackage{numbers, compress}{natbib}

\usepackage[preprint]{neurips_2026}

\usepackage[utf8]{inputenc} 
\usepackage[T1]{fontenc}   
\usepackage[colorlinks=true,citecolor=purple,linkcolor=purple]{hyperref}     
\usepackage{url}           
\usepackage{booktabs}      
\usepackage{amsfonts}      
\usepackage{nicefrac}       
\usepackage{microtype}      
\usepackage{xcolor}

\usepackage{amsthm}
\usepackage{graphicx}
\usepackage{wrapfig}
\usepackage{multicol}
\usepackage{multirow}
\usepackage{diagbox}

\theoremstyle{definition}
\newtheorem{definition}{Definition}[section]

\newcommand{\GRAFO}{\underline{Gra}ph-based \underline{Fo}rmation (GRAFO)}
\newcommand{\FROG}{\underline{Fro}zen \underline{G}raph Encoder (FROG)}

\usepackage{amsmath, amsfonts, bm}

\usepackage{etoolbox}
\makeatletter
\patchcmd{\hyper@makecurrent}{%
    \ifx\Hy@param\Hy@chapterstring
        \let\Hy@param\Hy@chapapp
    \fi
}{%
    \iftoggle{inappendix}{
        \@checkappendixparam{chapter}%
        \@checkappendixparam{section}%
        \@checkappendixparam{subsection}%
        \@checkappendixparam{subsubsection}%
        \@checkappendixparam{paragraph}%
        \@checkappendixparam{subparagraph}%
    }{}%
}{}{\errmessage{failed to patch}}

\newcommand*{\@checkappendixparam}[1]{%
    \def\@checkappendixparamtmp{#1}%
    \ifx\Hy@param\@checkappendixparamtmp
        \let\Hy@param\Hy@appendixstring
    \fi
}
\makeatletter

\newtoggle{inappendix}
\togglefalse{inappendix}

\apptocmd{\appendix}{\toggletrue{inappendix}}{}{\errmessage{failed to patch}}

\def\eqref#1{equation~\ref{#1}}

\def\1{\bm{1}}

\def\va{{\bm{a}}}

\def\ve{{\bm{e}}}

\def\vg{{\bm{g}}}

\def\vo{{\bm{o}}}

\def\vs{{\bm{s}}}

\def\vx{{\bm{x}}}

\def\mE{{\bm{E}}}

\def\mX{{\bm{X}}}

\DeclareMathAlphabet{\mathsfit}{\encodingdefault}{\sfdefault}{m}{sl}
\SetMathAlphabet{\mathsfit}{bold}{\encodingdefault}{\sfdefault}{bx}{n}

\def\gA{{\mathcal{A}}}
\def\gB{{\mathcal{B}}}

\def\gE{{\mathcal{E}}}

\def\gG{{\mathcal{G}}}

\def\gK{{\mathcal{K}}}

\def\gM{{\mathcal{M}}}
\def\gN{{\mathcal{N}}}
\def\gO{{\mathcal{O}}}
\def\gP{{\mathcal{P}}}

\def\gR{{\mathcal{R}}}
\def\gS{{\mathcal{S}}}
\def\gT{{\mathcal{T}}}

\def\gZ{{\mathcal{Z}}}

\def\sR{{\mathbb{R}}}

\DeclareMathOperator*{\aggr}{\textsc{Aggr}}

\usepackage[acronym, nohypertypes={acronym}]{glossaries}

\newacronym{gnn}{GNN}{Graph Neural Network}

\title{Continual Graph Multi-Agent\\
Reinforcement Learning}

\author{Tommaso Marzi\textsuperscript{\rm 1},
Ahmed Hendawy\textsuperscript{\rm 2,3},
Jan Peters\textsuperscript{\rm 2,3,4,5},
Carlo D'Eramo\textsuperscript{\rm 6}\\ \textbf{Andrea Cini}\textsuperscript{\rm 7}, \textbf{Cesare Alippi}\textsuperscript{\rm 1,8}\\
\textsuperscript{\rm 1}IDSIA USI-SUPSI, Università della Svizzera italiana, Lugano, Switzerland\\
\textsuperscript{\rm 2}Technical University of Darmstadt, Darmstadt, Germany\\
\textsuperscript{\rm 3}Hessian.AI\\
\textsuperscript{\rm 4}Robotics Institute Germany (RIG)\\ \textsuperscript{\rm 5}German Research Center for AI (DFKI)\\
\textsuperscript{\rm 6}University of Würzburg, Würzburg, Germany\\
\textsuperscript{\rm 7} EPFL, Lausanne, Switzerland\\
\textsuperscript{\rm 8} Politecnico di Milano, Milan, Italy
}

\begin{document}

\maketitle

\begin{abstract}
In Continual Multi-Agent Reinforcement Learning (CMARL), agents learn cooperative policies across sequences of tasks, aiming to adapt effectively to new tasks while preserving the ability to solve previously encountered ones. In many applications, tasks differ in their underlying \textit{structure}, which can represent, for example, distinct operational conditions or target configurations~(e.g., different network topologies in power grids or arrangements in formation control). 
Existing CMARL methods lack dedicated mechanisms to leverage this structural information when learning new tasks, failing to promote transfer and mitigate forgetting. 
To fill this gap, we propose Continual Graph Multi-Agent Reinforcement Learning (CGMARL), a novel framework for CMARL problems in which task sequences are mapped into a series of attributed graphs, each modeling a task-specific structure. In CGMARL, each graph determines the environment dynamics (next states and/or rewards) and the number of agents for the corresponding task. 
Then, we present \GRAFO, the first CGMARL benchmark, and show how forgetting arises in this setting. Finally, to address this limitation, we propose \FROG, a method that relies on a frozen graph backbone to preserve past structural information in graph-based CMARL policies. Experiments on GRAFO show that pairing FROG with existing CL methods substantially improves performance on multiple CGMARL scenarios.
\end{abstract}

\section{Introduction}
In cooperative Multi-Agent Reinforcement Learning~(MARL;~\citep{littman1994markov}), multiple RL agents learn to coordinate their behaviors to accomplish a shared task. This paradigm has proven effective across a variety of domains, such as autonomous driving~\citep{candela2022transferring}, traffic signal control~\citep{kolat2023multi}, or video games~\citep{vinyals2019grandmaster}. 
However, its applicability remains limited: most existing methods train RL agents on a single problem, whereas real-world scenarios typically involve sequences of related tasks that are often not known in advance.
Continual RL~(CRL;~\citep{khetarpal2022towards,abel2023definition}) formalizes this challenge by considering RL environments in which the agent never stops learning from an unknown sequence of tasks and cannot access previously observed ones. In particular, the agent aims to adapt to new tasks while preserving performance on previously learned ones~\citep{mccloskey1989catastrophic,parisi2019continual}, i.e., to transfer knowledge across tasks (\textit{plasticity}) while avoiding catastrophic forgetting (\textit{stability}). 
The multi-agent extension of the CRL paradigm, namely, Continual MARL~(CMARL;~\citep{pan2025survey,tomilin2026meal}), poses additional challenges. Specifically, learning new tasks while mitigating the forgetting of past ones is complicated by multiple interacting learners~\citep{yuan2024multiagent}. This introduces a tightly coupled adaptation problem: agents must adapt to external changes in the environment, while accounting for changes in other agents' behaviors due to policy updates.

In many real-world domains, such as autonomous mobility-on-demand~\citep{gammelli2021graph} or power grid control~\citep{yoon2021winning,hassouna2026graph}, the multi-agent problem is characterized by an underlying \textit{structure}~\citep{liu2025survey}, which encodes dependencies across agents (e.g., physical or relational).
Importantly, this structure determines the environment's dynamics, i.e., it shapes state transitions and rewards. For example, in a traffic light network where each agent controls a specific intersection, local rewards and observations depend on incoming/outgoing vehicles from/to \textit{connected} agents.
Crucially, agents often face sequences of MARL tasks that differ in their underlying structures~\citep{chen2026multi}, requiring them to adapt their strategy to distinct structure-dependent dynamics. These structures may be entirely unrelated, such as flying formation problems with distinct target configurations, or may share similarities, e.g., adding or removing intersections and roads in a traffic light network. 
However, the current CMARL framework is not suited to address such sequential structure-based MARL problems: task shifts are typically attributed to variations in environment dynamics or team size~\citep{pan2025survey}, whereas the impact of task-specific problem structures is largely overlooked. 
Moreover, standard CL methods lack dedicated mechanisms to model \textit{structural} information~\citep{zhou2021overcoming}, making them ineffective at preserving past structural knowledge when adapting to new tasks. For example, regularization-based methods~\citep{EWC} mitigate forgetting by preserving important features or parameters, but fundamentally ignore the task's structure. 
Given these limitations, we argue that \textit{graphs} provide a natural and effective approach for capturing task-specific structures by representing agents as nodes and encoding pairwise (agent-agent) dependencies through edges.

In this work, we introduce Continual Graph Multi-Agent Reinforcement Learning (CGMARL), a novel framework for continual structure-based MARL problems. In CGMARL, task sequences are characterized by a series of attributed graphs, each modeling a task-specific structure and determining the environment dynamics and the number of agents. 
We first show how distinct graph distributions induce CGMARL problems relevant to real-world scenarios. Then, we present the first CGMARL benchmark, named \GRAFO, and show the failure modes of graph-based policies in CGMARL.
Finally, we propose \FROG, a simple approach tailored for CGMARL problems that relies on decoupling graph representation learning from action prediction. Specifically, FROG leverages node representations generated with a frozen graph backbone to \textit{provably} eliminate forgetting in the graph representation learning module caused by structural changes.   
At the same time, the action prediction module can be paired with standard CL approaches to further mitigate forgetting. 
To the best of our knowledge, this is the first work that formalizes continual graph-based MARL problems and can pave the way for novel methodologies tailored to CGMARL. 

\textbf{Main Contributions:} \textbf{1)} We introduce the Continual Graph Multi-Agent Reinforcement Learning (CGMARL) framework, in which continual structure-based MARL problems are modeled as a series of attributed graphs. \textbf{2)} We present \GRAFO, a novel graph-based navigation benchmark that allows us to design arbitrary CGMARL task sequences. 
\textbf{3)} We provide empirical evidence that, in CGMARL, standard graph-based methods fail to generalize across different graphs and strongly suffer from forgetting. 
\textbf{4)} We propose \FROG, a simple technique that relies on a frozen graph backbone, and showcase its effectiveness across multiple CGMARL sequences. 

\section{Preliminaries}~\label{sec:preliminaries}
\textbf{Cooperative MARL:} 
Cooperative MARL problems can be modeled as a Decentralized Partially Observable Markov Decision Process~(Dec-POMDP;~\citep{oliehoek2016concise}), i.e., as a tuple: 
\begin{equation}
  \gM = \langle \gN, \gS, \bm{\gO} = \{\gO^i\}_{i\in\gN}, \bm{\gA} =\{\gA^i\}_{i\in\gN}, \mathcal{P},\gR, \gamma \rangle  
  \label{eq:dec_pomdp}
\end{equation}
where $\gN$ is the agent set, $\gS$ is the state space, $\gO_i$ and $\gA_i$ are the $i$-th agent's observation and action spaces, $\mathcal{P}: \gS \times \bm{\gA}\rightarrow \gS$ and $\gR: \gS \times \bm{\gA}\rightarrow \sR$ are the transition and reward functions depending on the joint state and action, and $\gamma\in[0,1)$ is a discount factor. Local observations $\{\vo_t^i\}_{i\in\gN}$ are given by an unknown function of the global state $\vs_t$. The agents' objective is to learn a joint policy $\boldsymbol{\pi}$ composed of agent-specific policies $ \{\pi^{i}\}_{i\in\gN}$ that maximizes the expected discounted cumulative reward, denoted as $J(\boldsymbol{\pi})$. We assume agents to be homogeneous~\citep{albrecht2024multi}, and the joint policy is a collection of identical single-agent policies, i.e., $\pi^i = \pi\;\forall\; i\in\gN$.

\textbf{Continual MARL:} In Continual MARL (CMARL; \cite{pan2025survey,tomilin2026meal}), the multi-agent system undergoes a (possibly infinite) sequence of tasks $\gT = \gM_1,\dots,\gM_K$, where each $\gM_k$ is a Dec-POMDP (\autoref{eq:dec_pomdp}) occurring for a fixed number of iterations. Tasks are drawn from a fixed, but unknown, task distribution~\citep{abel2018policy}. The subscript $k$ denotes a Dec-POMDP with potentially different components, i.e., different $\gN_k, \mathcal{P}_k,\gR_k$.\footnote{The most general formulation of CRL~\citep{khetarpal2022towards} allows for task-dependent state, observation, and action spaces, and the same can be adopted for Continual MARL. However, in practice, they are assumed to be fixed to facilitate learning~\citep{pan2025survey}.}
In the remainder, we denote tasks using the index $k$ to distinguish them from the time step $t$ within an RL episode.

\section{Continual Graph MARL}~\label{sec:cgmarl}
In this section, we formalize the Continual Graph MARL (CGMARL) framework. First, we introduce the graph-based MARL problem, which corresponds to the single-task formulation. Then, after defining the CGMARL problem, we illustrate relevant task sequences.

\subsection{Graph-based MARL Task}~\label{sec:single_task}
In a graph-based MARL problem, the Dec-POMDP in~\autoref{eq:dec_pomdp} is further equipped with a graph $\gG = \langle\gN,\gE,\mX,\mE\rangle$, with $\gN$ and $\gE$ representing the set of nodes~(agents) and edges, respectively, each equipped with feature matrices $\mX\in\mathbb{R}^{|\gN|\times d_n}$ (nodes) and $\mE\in\mathbb{R}^{|\gE|\times d_e}$ (edges).
Importantly, the graph $\gG$ represents the task structure and determines the environment dynamics, i.e., the transition function $\mathcal{P}_{\gG}$ and/or reward function $\gR_{\gG}$ depend on $\gG$, providing local observations and rewards~\citep{schneider1999distributed}. Thus, the Dec-POMDP tuple in~\autoref{eq:dec_pomdp} becomes:
\begin{equation}
\label{eq:nmarl}
\gM = \langle \gG, \gS, \bm{\gO}, \bm{\gA}, \mathcal{P}_{\gG},\gR_{\gG}, \gamma \rangle
\end{equation}
For example, in a traffic light network, rewards and next observations depend on incoming and outgoing vehicles at each intersection, i.e., the $1$-hop neighbors of each node in $\gG$. Furthermore, agents can propagate messages through the network to coordinate effectively.
We note that $\mX$ and $\mE$ may contain both static and dynamic features.
In graph-based MARL, dynamic features are given by the agents' observations, which change during the episode. Therefore, we use $\mX$ and $\mE$ to denote the \textit{static graph features} only, representing fixed node (e.g., agent identifier or type) and edge (e.g., sensor distances) features, respectively.
We represent the edge set $\gE$ with the adjacency matrix $A\in\{0,1\}^{|\gN|\times |\gN|}$, where $a_{ij}=1$ if an edge connects node $j$ to $i$. 
In the general formulation, $A$ can vary within a task, i.e., the edge set $\gE$ can change across steps (e.g., in navigation problems with distance-based connections). 
Specifically, edges are given by a state-dependent connectivity function $\varphi: \gN\times \gS \to \{0,1\}^{|\gN|\times |\gN|}$, which, at each step $t$, generates the adjacency matrix of the graph according to the current state $\vs_t$. In this case, the graph is characterized by the tuple $\gG = \langle \gN, \varphi, \mX, \mE\rangle$, where we remind that the non-static graph features consist of the observations.

\textbf{Comparison with Communication MARL:} Graph-based MARL is fundamentally different from communication MARL~\citep{zhu2024survey}. In our formulation, the graph represents the problem structure and determines the dynamics. This is more similar to other graph-based MDPs, where, e.g., each node's rewards and observations depend on its neighbors in the graph~\citep{hu2024decentralized,rashwan2026factored}. Conversely, communication MARL focuses on developing effective communication methods across agents rather than using graphs to represent environment configurations with different dynamics. Specifically, the transition $\mathcal{P}$ and reward $\gR$ functions do not depend on the graph structure $\gG$, which is used only for message passing. 
While communication methods can exchange messages on different graph structures, they assume the underlying task to be fixed and, as a result, fail to adapt to unseen problems.  
We empirically analyse these failure modes in~\autoref{sec:forgetting_cgmarl}. 

\subsection{CGMARL Sequence}
We now formalize a Continual MARL problem in which the task sequence can be represented by a series of attributed graphs, each defining a specific graph-based MARL task (\autoref{eq:nmarl}). We assume the graph-dependent environment dynamics (i.e., $\gP_{\gG}$ and $\gR_{\gG}$) to be fixed across tasks. Thus, each task $\gM_k$ in the sequence is uniquely identified by the graph $\gG_k$, which determines the transition function $\gP_{\gG_k}$ and reward function $\gR_{\gG_k}$. We now define the Continual Graph MARL~(CGMARL) problem.

\begin{definition}[CGMARL Problem]~\label{def:cgmarl}
Let $\gS, \bm{\gO}, \bm{\gA}, \gamma $ be the fixed components of the graph-based Dec-POMDP~(\autoref{eq:nmarl}) representing state, observation, and action spaces, and discount factor, respectively. Let $\gG= \langle \gN, \varphi, \mX, \mE\rangle$ be a graph, and let $\gP_{\gG}$ and $\gR_{\gG} $ be the graph-dependent transition and reward functions. Given a fixed but unknown distribution $\gZ$ over graphs, the Continual Graph Multi-Agent Reinforcement Learning~(CGMARL) problem is defined as a sequence of tasks $\gT = \gM_1,\dots,\gM_K$, where each task $\gM_k$ is defined as in~\autoref{eq:nmarl} and characterized by the graph $\gG_k$ sampled from the distribution $\gZ$.
\end{definition}

\begin{figure}[t]
\centering
\includegraphics[width=\textwidth]{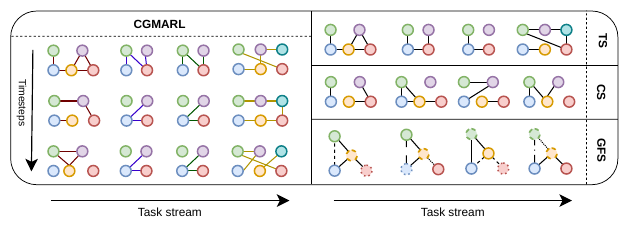}
\vspace{-1cm}
\caption{Scheme of the CGMARL problem. In the left panel, we illustrate the general setup, in which edges are generated dynamically across timesteps using task-specific connectivity functions (indicated by different colors for edges). In the right panel, we illustrate relevant classes of the CGMARL problem. In Topology Switch (TS), both nodes and edges can change across tasks. In Connectivity Switch (CS), edges can change while the node set is fixed. In Graph Features Switch (GFS), the topology is kept static, and graph features can change; this is highlighted by using different dashed patterns to represent nodes and/or edges.}
\label{fig:problem_setting}
\end{figure}

This is the most general formulation of Continual Graph MARL. This class of problems can be motivated by considering a navigation task with distance-based connections. In this setting, each pair of agents is connected if their distance is within a specified threshold, such as the vision range~\citep{nayak2023scalable}. Across tasks, the team composition $\gN_k$ may vary depending on, e.g., the number of target positions to be covered or map size. Additionally, different scenarios may exhibit varying thresholds, thereby influencing the connectivity function $\varphi_k$ that generates the communication structure (edge set) at each time step. Depending on the graph distribution $\gZ$, we can derive different classes of CGMARL problems, each with its relevance and potential applicability. In the following, we present formal definitions alongside real-world examples fitting the CGMARL framework in decreasing order of generality. Please refer to~\autoref{fig:problem_setting} for a visual summary.

\subsubsection{Topology Switch}~\label{sec:ts_scenario}
In the Topology Switch (TS) problem, connections (edge set $\gE_k$) remain fixed \textit{within} a task, but may vary across \textit{different} tasks. Formally, for each task $\gM_k$, the graph $\gG_k$ sampled by the distribution $\gZ$ has a connectivity function $\varphi_k$ that is state-independent, thus generating a fixed edge set $\gE_k=\varphi_k(\gN_k)$. In other words, the adjacency matrix $A_k$ remains fixed within the task $\gM_k$.

\begin{definition}[TS Problem]~\label{def:ts_problem}
We call the Topology Switch (TS) problem a CGMARL problem where the task sequence $\gT$ is generated under state-independent connectivity functions $\varphi: \gN \to \{0,1\}^{|\gN|\times |\gN|}$; it follows that each task $\gM_k$ of $\gT$ is characterized by a graph $\gG_k = \langle\gN_k, \gE_k, {\mX}_k, {\mE}_k\rangle $ with a static edge set $\gE_k = \varphi_k(\gN_k)$.
\end{definition}

A graph-based control policy for efficient traffic light control~\citep{8569301,9405489} is an example of a TS problem. In this setting, the same model could be deployed and updated on completely different network topologies, e.g., different cities or districts within a city.  
Importantly, the connections do not change within the same task.

\subsubsection{Connectivity Switch}~\label{sec:cs_scenario}
The Connectivity Switch (CS) problem is a sub-case of the TS setting in which the node set remains fixed across tasks. In other words, in this CGMARL setting, we consider a \textit{static} team of agents that cooperate to solve different tasks, each with its own communication structure and graph features. Therefore, the sampling process from $\gZ$ corresponds to sampling a sequence of different adjacency matrices\footnote{With a state-dependent connectivity function, edges could also change \textit{within} a single task. However, unless the environment dynamics vary, the fixed node set restricts the possible edge configurations, inducing task similarity. We leave this variant of the CS setting to future work.} and graph features given a fixed node set.

\begin{definition}[CS Problem]
We call the Connectivity Switch (CS) problem a CGMARL problem in which the task sequence $\gT_{{\gN}}$ is generated under a specific choice for the node set ${{\gN}}$; it follows that each task $\gM_k$ of $\gT_{\gN}$ is characterized by a graph $\gG_k = \langle{\gN}, \gE_k, {\mX}_k, {\mE}_k\rangle $, with $\gE_k=\varphi_k(\gN)$.
\end{definition}

The CS problem can be relevant, for example, in efficient power distribution management~\citep{yoon2021winning}. Here, the physical network representing the power grid consists of a fixed set of nodes (e.g., substations), while the connections may vary according to the current operational regime. Specifically, the network can transition between predefined edge configurations in response to external operational conditions (e.g., fault isolation or maintenance). 

\subsubsection{Graph Features Switch}~\label{sec:gfs_scenario}
When the topology (i.e., node and edge sets) remains fixed throughout the task sequence, static graph features can discriminate between different graphs~(tasks). In the Graph Features Switch (GFS) problem, the graph distribution $\gZ$ is a distribution over static graph features ($\mX$ and $\mE$). 

\begin{definition}[GFS Problem]
We call the Graph Features Switch (GFS) problem a CGMARL problem in which the task sequence $\gT_{{\gN},{\gE}}$ is generated under a specific choice for the topology (${\gN}$ and ${\gE}=\varphi(\gN)$); it follows that each task $\gM_k$ of $\gT_{\gN,\gE}$ is characterized by a graph $\gG_k = \langle{\gN}, {\gE}, {\mX}_k, {\mE}_k\rangle $.
\end{definition}

In this class of CGMARL problems, graph features play a critical role in tackling the task. For instance, consider a robotic agent where each joint~(actuator) is represented as a node, and the limbs connecting them are the edges~\citep{wang2018nervenet,marzi2024feudal,yang2024transfer}. In scenarios where limb lengths depend on the task, it is essential to learn a policy that adapts to different configurations. For example, limbs may be lengthened or shortened in response to different obstacles or terrain. In this case, ${\mE}_k$ can encode task-dependent distances between node pairs, allowing robots to operate effectively with longer or shorter limbs. Similarly, node features ${\mX}_k$ can encode information on task-specific conditions, such as node-level body masses.  
Note that, unlike TS and CS sequences where the topology implicitly serves as the communication structure, in the GFS problem the features $\mX$ and $\mE$ characterizing the graph are \textit{explicitly} used in decision-making.

\begin{wrapfigure}{r}{0.5\textwidth}
\centering
\vspace{-0.55cm}
\includegraphics[width=0.5\textwidth]{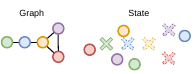}
\vspace{-0.7cm}
\caption{Example of a GRAFO task with five agents (circles). Left: agent graph identifying the task. Right: current state of the environment, where the green solid cross is the landmark known to all agents, i.e., the target position of the green agent, while the dashed crosses are the target positions of other agents. These positions are defined based on the graph, but are not explicitly provided.}
\label{fig:env_graph}
\end{wrapfigure}

\section{GRAFO Benchmark}~\label{sec:grafo_benchmark}
As a first testbed for CGMARL problems, we propose \GRAFO, a novel graph-based navigation benchmark that supports generating arbitrarily long CGMARL sequences. In GRAFO, the agents' objective is to navigate to a target configuration determined by the static input graph characterizing the task (refer to~\autoref{fig:env_graph} for a visual example). Specifically, a landmark (the green, solid cross in~\autoref{fig:env_graph}) represents the target position of a specific agent (the green circle), and it is known to all agents. Instead, other agents' targets (dashed crosses) are not directly observable and are implicitly defined by the graph topology. Since only one target is explicitly provided, agents must leverage structural information to identify the correct formation. Each agent must reach its target position, and the agents' graph-based rewards are dense signals based on the distances from their assigned targets, i.e., they depend on the task-specific graph~(\autoref{sec:single_task}). CGMARL sequences can be easily defined by considering a series of graphs~(\autoref{sec:task_sequences}).
We report a more detailed description of GRAFO in~\autoref{app:cgmarl_environment}.

\section{Emergence of Forgetting in CGMARL}

In this section, we first show how GNN-based policies fail to preserve structural information in CGMARL problems. Then, we present a simple method to mitigate forgetting in this setting.

\subsection{GNN-based Policies in Graph-based MARL}~\label{sec:forgetting_cgmarl}
Graph Neural Networks~(GNNs;~\citep{bacciu2020gentle}) operate on graph-structured data and generate node-level representations by aggregating information over each node's neighborhood. The updated representation $\vx_i'$ of a node $i$ with neighboring nodes $\mathcal{B}(i)$ can be written as~\citep{gilmer2017neural}:
\begin{equation}
    \vx_i' = \psi_U\Big(\vx_{i}, \aggr_{j \in \mathcal{B}(i)}\Big\{\psi_M\big(\vx_{i}, \vx_{j}; \ve_{ij}\big)\Big\}\Big),
    \label{eq:mp}
\end{equation}
where $\ve_{ij}\in\mathbb{R}^{d_e}$ is the feature vector of edge $a_{ij}$, $\aggr$ is a permutation-invariant aggregation function, and $\psi_U$, $\psi_M$ are learnable update and message functions, respectively. GNNs are designed to process graphs of \textit{any} size, i.e., with a varying number of nodes and edges. This makes GNN-based policies suitable candidates to generalize naturally across different graph-based MARL tasks. However, we argue that this cannot be easily achieved: GNNs learn local (node-level) rules that may not directly apply to unseen tasks if the underlying graph-dependent dynamics change. When considering CGMARL sequences, this requires overwriting previously learned rules when adapting to new tasks, making GNN-based policies prone to forgetting.

\def\thickhline{\noalign{\hrule height1.3pt}}

\begin{wraptable}[14]{R}{0.5\textwidth}
\vspace{-0.6cm}
\caption{Solved percentage at test time (median $\pm$ standard deviation of $5$ seeds) for two GRAFO tasks in both zero-shot transfer (TRAN.) and continual (CONT.) settings. }
\vspace{0.2cm}
\label{tab:zero_shot_performance}
\renewcommand{\arraystretch}{1.5}
\centering
\resizebox{0.5\columnwidth}{!}{
\begin{tabular}{c|c|c|c}
\thickhline
&\diagbox[width=2.5cm, height=0.75cm]{Train}{Test} & GRAFO-$3$ & {GRAFO}-$5$ \\[1pt]
\thickhline
 \parbox[t]{2mm}{\multirow{2}{*}{\rotatebox[origin=c]{90}{\footnotesize{TRAN.}}}} & GRAFO-$3$ &  $\boldsymbol{1}\pm \scriptstyle{0}$ & ${0}\pm \scriptstyle{0}$ \\
& GRAFO-$5$&  ${0.016}\pm \scriptstyle{0.016}$ & $\boldsymbol{1}\pm \scriptstyle{0}$ \\
\hline
\parbox[t]{2mm}{\multirow{2}{*}{\rotatebox[origin=c]{90}{\footnotesize{CONT.}}}} & GRAFO-$(3, 5)$&  ${0.000}\pm \scriptstyle{0.027}$ & $\boldsymbol{1}\pm \scriptstyle{0}$\\
& GRAFO-$(5, 3)$&  $\boldsymbol{1}\pm \scriptstyle{0}$&  ${0.000}\pm \scriptstyle{0.014}$  \\
\thickhline
\end{tabular}
}
\end{wraptable}
To empirically assess these observations, we evaluate InforMARL~(\autoref{app:informarl};~\citep{nayak2023scalable}) on two GRAFO tasks, from both zero-shot and continual perspectives. Specifically, we train the model on graphs with $3$ and $5$ nodes individually (GRAFO-$3$ and GRAFO-$5$) and sequentially (GRAFO-$(3, 5)$ and GRAFO-$(5,3)$), and test it on both tasks (further details are reported in~\autoref{sec:task_sequences}). The results reported in~\autoref{tab:zero_shot_performance} support our claim. In particular, the zero-shot transfer (TRAN.) results show that the model solves all tasks individually, but the learned behavior cannot be directly transferred to different target formations. When learning sequentially (CONT.), the model fails to retain previously acquired knowledge, resulting in complete forgetting of the first task learned (e.g., task GRAFO-$3$ in the GRAFO-$(3,5)$ sequence).

We did not apply any CL strategy (e.g., feature importance,~\citep{EWC}) in the sequential (CONT.) setting of~\autoref{tab:zero_shot_performance}. 
Standard CL methods can mitigate forgetting but, as also shown by the results in \autoref{sec:experiments}, they do not explicitly account for changes encoded in the graph representation of the task. In the following, we propose a simple method specifically tailored to CGMARL problems.

\subsection{FROG: A Simple Approach to Mitigate Forgetting in CGMARL }~\label{sec:FROG}
Standard CL approaches do not account for forgetting in graph representation learning. We address this by decoupling graph processing from action prediction.  Specifically, consider a policy $\pi_{\boldsymbol{\theta}} = \mu_{\theta_2} \circ\psi_{\theta_1}$, where $\psi_{\theta_1}$ is a generic GNN as in~\autoref{eq:mp} with learnable parameters $\theta_1$ and $\mu_{\theta_2}$ is an action prediction module (e.g., MLP) with learnable parameters $\theta_2$. Actions are taken as:
\begin{equation}
\label{eq:grf_policy}
    \va_t^i \doteq \pi_{\boldsymbol{\theta}}(\vo_t^i;\gG) = \mu_{\theta_2}(\psi_{\theta_1}(\vo_t^i;\gG)),
\end{equation}
where we assumed a deterministic policy~(it can be easily generalized to a stochastic one). 
Under this formulation, the graph encoder $\psi_{\theta_1}$ captures the structural information, while the action module $\mu_{\theta_2}$ directly maps node representations to actions. This facilitates analyzing forgetting associated with graph representation learning.  
Formally, given task indexes $k$ and $h$, with $k>h$, forgetting on task $h$ after training on task $k$ is defined as (details in~\autoref{app:theory_forgetting}):
\begin{equation}
\label{eq:forgetting}
   J_h(\boldsymbol{\pi}_k) - J_h(\boldsymbol{\pi}_h) \leq
   \underbrace{\lVert\boldsymbol{\mu}_k(\boldsymbol{\psi}_k(\vo;\gG_h)) - \boldsymbol{\mu}_k(\boldsymbol{\psi}_h(\vo;\gG_h))\rVert_\infty}_{\text{Graph Representation Learning}} 
   +\underbrace{\lVert\boldsymbol{\mu}_k(\boldsymbol{\psi}_h(\vo;\gG_h)) - \boldsymbol{\mu}_h(\boldsymbol{\psi}_h(\vo;\gG_h))\rVert_\infty}_{\text{Action Prediction}}  
\end{equation}
where $J_i(\boldsymbol{\pi}_j)$ is the performance (e.g., expected return) achieved on task $i$ using joint policy $\boldsymbol{\pi}_j$ (with parameters $j:= \boldsymbol{\theta}_j=\{\theta_{1,j},\theta_{2,j}\}$ after training task $j$).
From the first term on the right-hand side of~\autoref{eq:forgetting}, we can \textit{provably} eliminate forgetting in the graph representation learning module caused by structural changes by imposing $\psi_k = \psi_h $ for each task pair $h,k$. This can be achieved by freezing the graph encoder $\psi$ either \textit{before} or \textit{after} training the first task. 
In our experiments, we adopt the second option and call our method \FROG. Specifically, freezing the graph encoder \textit{after} learning the first task preserves its learned parameters while allowing it to generate distinct graph representations for each task-specific graph. These are then passed to the action module $\mu_{\theta_2}$, which continuously adapts to new tasks. To mitigate forgetting in this module~(second term on the right-hand side of~\autoref{eq:forgetting}), we pair it with existing CL methods. In this way, we leverage the generalization capacity of the GNNs~(\autoref{sec:forgetting_cgmarl}) while confining task-specific adaptation to the action prediction module. Note that freezing the graph encoder \textit{before} training provides node-level representations as random graph features~\citep{zambon2020graph, navarin2023untrained}, which have been proven effective in CL problems on graphs~\citep{donghi2025unreasonable}. Nonetheless, we argue that this option may hinder learning in complex graph-based MARL settings, especially if the GNN lacks sufficient expressive power.
We compare FROG with this untrained variant in~\autoref{sec:ablation}.

\section{Empirical Evaluation}~\label{sec:experiments}
In this section, we present a series of experiments to empirically support our claims. Specifically, we have two main objectives: evaluate how standard CL methods perform in CGMARL problems (\textbf{Obj.~1}), and investigate whether FROG, i.e., a simple method designed to account for structural changes, can provide benefit in this setting~(\textbf{Obj.~2}), paving the way for novel CGMARL approaches. 

\textbf{CGMARL Sequences:} We consider multiple CGMARL sequences from our proposed GRAFO benchmark (please refer to~\autoref{app:cgmarl_environment} and~\autoref{sec:task_sequences} for further details). Specifically, for the Topology Switch (TS) scenario (\autoref{sec:ts_scenario}), we consider a sequence of $6$ tasks, with $3$ to $7$ agents; the Connectivity Switch (CS) scenario (\autoref{sec:cs_scenario}) is a sequence of $5$ tasks, each with $6$ agents; for the Graph Feature Switch (GFS) scenario (\autoref{sec:gfs_scenario}), we consider a sequence of $4$ tasks, each with $4$ agents. Furthermore, we analyze the impact of task order for the TS sequence in~\autoref{sec:ablation}. Following related work~\citep{tomilin2026meal}, we consider the task-incremental learning paradigm with separate task heads and known task identities. We assess the impact of conditioning on task identities in~\autoref{sec:task_id_ablation}.

\textbf{Models:}  We employ InforMARL~\citep{nayak2023scalable} as the single-task GNN-based policy, as it is a well-established model for graph-based navigation domains (more details in~\autoref{app:informarl}).
To address forgetting in the CGMARL sequence, we pair InforMARL with common CL approaches~\citep{tomilin2026meal}. 
Specifically, we consider L2, where we apply an L2 penalty to the weights, Elastic Weight Consolidation~(EWC;~\citep{EWC}), and Memory Aware Synapses~(MAS;~\citep{MAS}). For EWC and MAS, we consider two variants depending on how parameter importance is computed: the "multi" (M) versions aggregate the importance matrix from all tasks via summation, while the "online" (O) versions employ an online exponential-decay running average of the importance matrix from past tasks. We pair FROG with all the CL methods, and consider an additional Fine-Tuning (FT) baseline without any regularization. Hyperparameters for InforMARL and the CL methods are reported in~\autoref{app:implementation}. 

\textbf{Metrics:} 
We evaluate the models using metrics from standard CL literature~\citep{wolczyk2021continual,caccia2022new,sokar2025continual}, i.e., Average Final Performance (AFP), Average Forgetting (AF), and Average Plasticity (AP). To do so, for each task, we measure the percentage of solved episodes during evaluation, where an episode is considered solved if the agents reach the target configuration within the time limit. Please refer to~\autoref{app:metrics} for a more detailed description.

\begin{figure}[t]
\centering
\includegraphics[width=\textwidth]{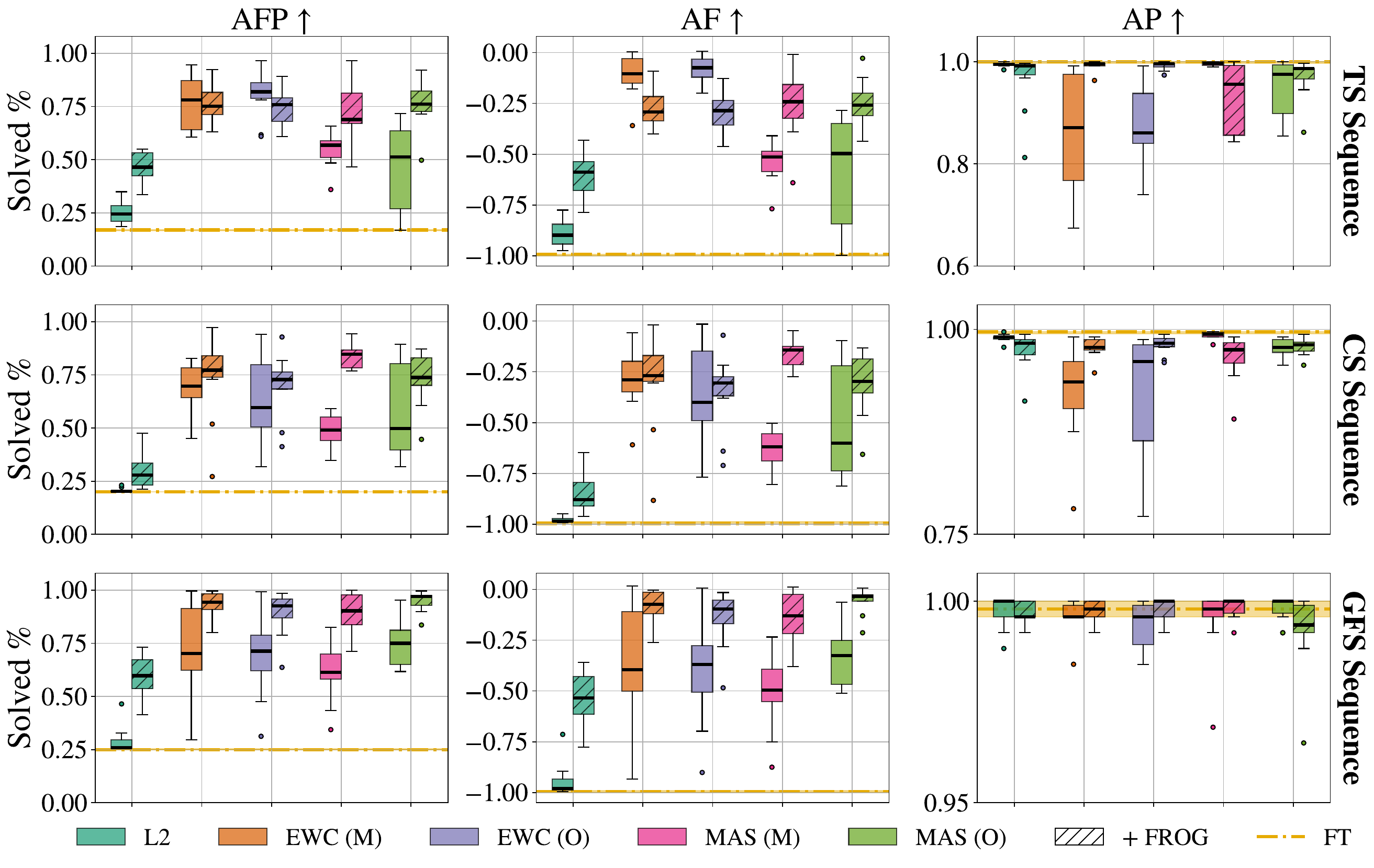}
\vspace{-0.7cm}
\caption{Boxplot showing the distribution of $10$ independent seeds for the different CL methods (with and without FROG) across the three scenarios, i.e., TS (top), CS (middle), and GFS (bottom). Boxes represent the IQR, horizontal lines indicate medians, and whiskers extend to $1.5\times\text{IQR}$.}
\label{fig:full_benchmark}
\end{figure}

\subsection{Main Results}~\label{sec:main_results}
We report the main results on three different CGMARL scenarios in~\autoref{fig:full_benchmark}. Standard CL methods exhibit mixed performance across the various sequences, with none consistently outperforming the others. On average, EWC achieves better AFP than MAS and L2, but at the cost of plasticity loss (lower AP). Combined with the high variance, this suggests that standard CL methods may require additional mechanisms tailored for CGMARL to be more effective (\textbf{Obj. 1}). Overall, compared with their counterparts that do not incorporate it, FROG consistently improves performance when combined with standard CL methods (\textbf{Obj. 2}), and is less prone to forgetting (higher AF). The only exception is EWC on the TS sequence, where the standard method has a higher AF value. However, this can be explained by plasticity loss: as the lower AP metric highlights, the standard version fails to learn new tasks. This reduced learning capacity necessarily leads to less forgetting, since the more difficult tasks are not successfully learned in the first place and therefore cannot subsequently be forgotten. We hypothesize that this may be caused by using a single regularization coefficient for both the graph encoder and action module, and we investigate this behavior in~\autoref{app:analysis_reg_coeff}. Finally, empirical evidence shows that, on average, using frozen graph representations stabilizes learning.

\subsection{Additional Analyses}~\label{sec:ablation}
In this section, we conduct a set of experiments to further assess the effectiveness of FROG.

\def\thickhline{\noalign{\hrule height1.3pt}}

\begin{wraptable}[8]{R}{0.5\textwidth}
\vspace{-0.6cm}
\caption{Aggregated medians (average $\pm$ standard deviation) for the different CL methods combined with FROG and FROG (RND) on the TS sequence. Complete boxplots are reported in~\autoref{fig:ablation_frog}. }
\vspace{0.2cm}
\label{tab:frog_analysis_untrained}
\renewcommand{\arraystretch}{1.1}
\centering
\resizebox{0.5\columnwidth}{!}{
\begin{tabular}{c|c|c|c}
\thickhline
 & AFP $\uparrow$ & AF $\uparrow$& AP $\uparrow$ \\
\thickhline
FROG &  $\boldsymbol{0.69}\pm \scriptstyle{0.13}$ & $-{0.33}\pm \scriptstyle{0.14}$& $\boldsymbol{0.99}\pm \scriptstyle{0.02}$\\
FROG (RND) &  ${0.38}\pm \scriptstyle{0.037}$ & $\boldsymbol{-0.29}\pm \scriptstyle{0.04}$& ${0.64}\pm \scriptstyle{0.03}$ \\
\hline
\end{tabular}
}
\end{wraptable}

\textbf{Untrained Graph Encoder:} We compare FROG with its untrained variant (FROG (RND)), where the graph encoder is frozen at initialization, and the corresponding representations are random. Aggregated results in~\autoref{tab:frog_analysis_untrained} (please refer to~\autoref{fig:ablation_frog} for a detailed plot) achieved in the TS sequence support our claim (\autoref{sec:FROG}): using completely random graph features hinders learning in GRAFO, as the model struggles to learn the hardest tasks (low AP).

\begin{figure}[t]
\centering
\includegraphics[width=\textwidth]{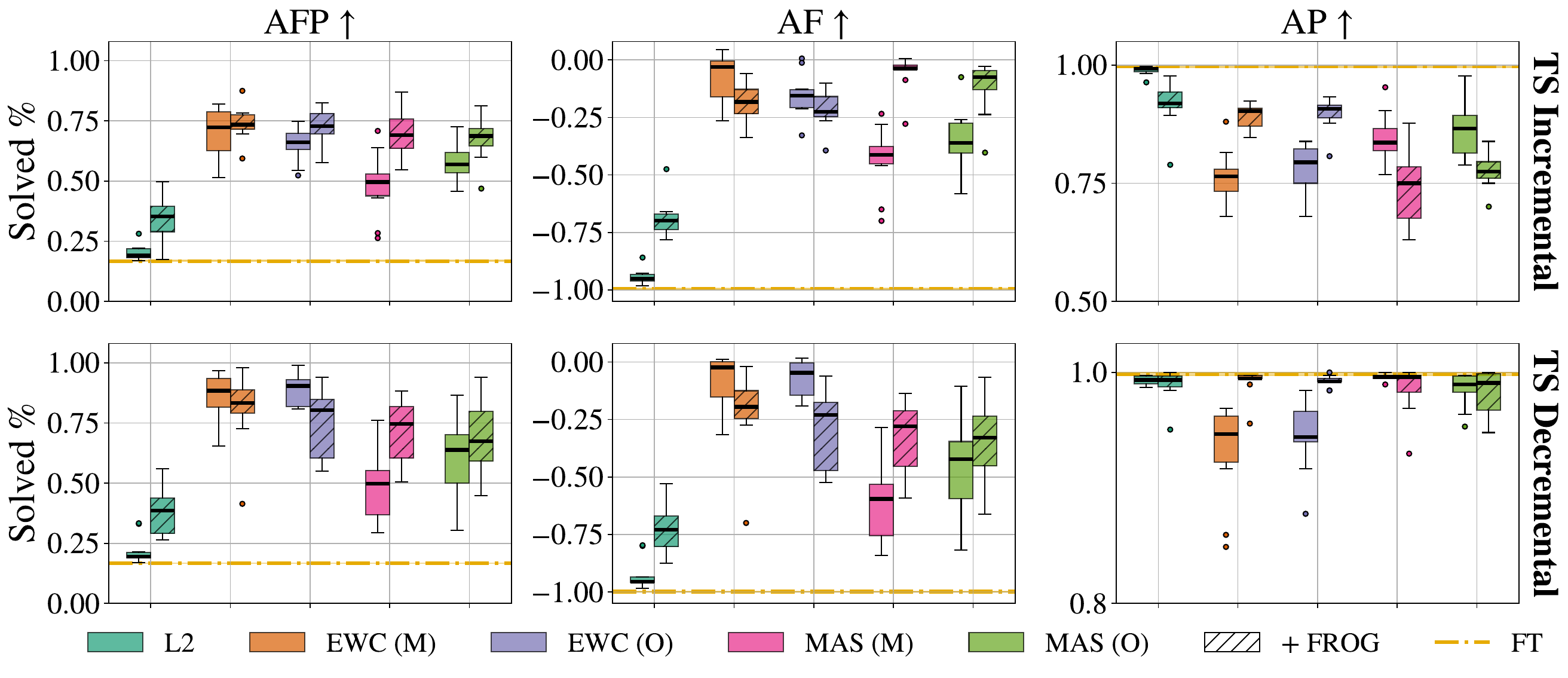}
\vspace{-0.7cm}
\caption{Boxplot showing the distribution of $10$ independent seeds for the different CL methods (with and without FROG) across the TS (inc) and TS (dec) sequences, i.e., the incremental/decremental variants of the TS sequence (top and bottom panels, respectively). Boxes represent the IQR, horizontal lines indicate medians, and whiskers extend to $1.5\times\text{IQR}$.}
\label{fig:TS_order}
\end{figure}

\textbf{Task Order:} FROG relies on freezing the graph encoder after training the first task. This implies that the quality of the node representations generated throughout the CGMARL sequence depends on the first task encountered.
To assess this behavior, we consider two variants of the TS sequence used in~\autoref{sec:main_results} with an incremental/decremental number of agents, and dub them TS (inc) and TS (dec), respectively. The results reported in~\autoref{fig:TS_order} show that, on average, FROG achieves competitive performance against the standard counterpart. 
In particular, in TS (inc), the first task requires the lowest coordination degree ($3$ agents). This causes plasticity loss (lower AP) in FROG when paired with MAS and L2, while it still guarantees good performance when combined with EWC. Conversely, in TS (dec), results are similar to the standard TS sequence, favoring FROG and suggesting that the learned graph encoder is rich enough to perform well on the full sequence. This shows that the first task can have an impact on FROG, but despite that, it can still achieve competitive performance. 
Finally, this analysis reveals that task order in GRAFO can also impact learning in standard CL methods, which achieve, on average, lower AFP and AP in the TS (inc) sequence.

\section{Related Work} 

Graph representations have been used in MARL to characterize the task structure in several settings. For example, many approaches~\citep{hu2024decentralized,rashwan2026factored} consider problems where local (agent-specific) dynamics are conditioned on actions of each agent's neighbors in the graph.
Practical applications of graph-based MARL range from networked systems~\citep{liu2025survey} to power grid control~\citep{hassouna2026graph}. 
Communication MARL~\citep{zhu2024survey} emerged as an effective paradigm for addressing those tasks by improving agents' coordination through communication. This is commonly achieved using Graph Neural Networks~(GNNs;~\citep{gilmer2017neural}) in decision-making to learn message-passing policies over the graph structure~\citep{Jiang2020Graph,niu2021multi, nayak2023scalable}.  
In Continual RL~(CRL;~\citep{khetarpal2022towards,abel2023definition}), the agent aims to sequentially adapt to new tasks without deteriorating performance on previously learned ones. To prevent forgetting, CL methods such as replay-based mechanisms~\citep{chaudhry2018efficient} or parameter regularizations~\citep{EWC} have been employed. The CRL problem has been extended to multi-agent systems~\citep{pan2025survey,tomilin2026meal}, but to the best of our knowledge, this is the first work that considers continual graph-based MARL settings. Related to our method, frozen modules or representations have been effectively used to mitigate forgetting in CRL~\citep{rusu2016progressive,powers2022self, nikishin2023deep,hu2026continual}. 
Please refer to~\autoref{app:ext_rw} for an extended discussion.

\section{Conclusion and Future Work}
In this work, we introduced the Continual Graph Multi-Agent Reinforcement Learning (CGMARL) framework, which formalizes sequences of MARL tasks characterized by different structures. Through this framework, we identified several real-world applications induced by different task~(graph) distributions. Motivated by the failure modes of standard approaches in CGMARL, we proposed \FROG, a simple approach that leverages frozen graph representations to mitigate forgetting. Experiments on a novel CGMARL benchmark, named \GRAFO, validated FROG's effectiveness when combined with standard CL methods.

We foresee several possible directions for future theoretical and empirical research. First, a theoretical investigation of the failure modes of standard CL approaches in this setting could help design more effective CGMARL methods. Furthermore, considering off-policy MARL algorithms would pave the way for testing experience-based CL methods. Lastly, it would be interesting to develop other CGMARL evaluation benchmarks, possibly with different practical applications (e.g., power distribution management or traffic light control).

\section{Acknowledgements}
TM and CA were supported by the Swiss National Science Foundation under project no.\ 204061 (High-Order Relations and Dynamics in Graph Neural Networks). AH was supported by the Deutsche Forschungsgemeinschaft (DFG, German Research Foundation) under Germany’s Excellence Strategy (EXC-3057/1 ’Reasonable Artificial Intelligence’, Project No. 533677015). AC was supported by the Swiss National Science Foundation under project no.\ 225351~(Relational Deep Learning for Reliable Time Series Forecasting at Scale). The authors would like to thank Giovanni Donghi for valuable discussions on continual learning and random graph features.

\clearpage

\bibliography{content}

@article{abel2023definition,
  title={A definition of continual reinforcement learning},
  author={Abel, David and Barreto, Andr{\'e} and Van Roy, Benjamin and Precup, Doina and van Hasselt, Hado P and Singh, Satinder},
  journal={Advances in Neural Information Processing Systems},
  volume={36},
  pages={50377--50407},
  year={2023}
}

@article{pan2025survey,
  title={A Survey of Continual Reinforcement Learning},
  author={Pan, Chaofan and Yang, Xin and Li, Yanhua and Wei, Wei and Li, Tianrui and An, Bo and Liang, Jiye},
  journal={arXiv preprint arXiv:2506.21872},
  year={2025}
}

@misc{tomilin2026meal,
title={{MEAL}: A Benchmark for Continual Multi-Agent Reinforcement Learning},
author={Tristan Tomilin and Luka van den Boogaard and Samuel Garcin and Constantin Ruhdorfer and Bram Grooten and Yali Du and Andreas Bulling and Mykola Pechenizkiy and Meng Fang},
year={2026},
url={https://openreview.net/forum?id=I3W8PynQU0}
}

@inproceedings{Chu2020Multi-agent,
title={Multi-agent Reinforcement Learning for Networked System Control},
author={Tianshu Chu and Sandeep Chinchali and Sachin Katti},
booktitle={International Conference on Learning Representations},
year={2020},
url={https://openreview.net/forum?id=Syx7A3NFvH}
}

@article{donghi2025online,
  title={Online Continual Graph Learning},
  author={Donghi, Giovanni and Pasa, Luca and Zambon, Daniele and Alippi, Cesare and Navarin, Nicol{\`o}},
  journal={arXiv preprint arXiv:2508.03283},
  year={2025}
}

@inproceedings{mallya2018packnet,
  title={Packnet: Adding multiple tasks to a single network by iterative pruning},
  author={Mallya, Arun and Lazebnik, Svetlana},
  booktitle={2018 IEEE/CVF Conference on Computer Vision and Pattern Recognition},
  pages={7765--7773},
  year={2018},
  organization={IEEE}
}

@article{donghi2025unreasonable,
author = {Donghi, Giovanni and Zambon, Daniele and Pasa, Luca and Alippi, Cesare and Navarin, Nicolò},
year = {2026},
month = {09},
pages = {},
title = {The Unreasonable Effectiveness of Randomized Representations in Online Continual Graph Learning},
volume = {115},
journal = {Machine Learning},
doi = {10.1007/s10994-026-07128-5}
}

@inproceedings{navarin2023untrained,
  title={An untrained neural model for fast and accurate graph classification},
  author={Navarin, Nicol{\`o} and Pasa, Luca and Gallicchio, Claudio and Sperduti, Alessandro},
  booktitle={International Conference on Artificial Neural Networks},
  pages={278--290},
  year={2023},
  organization={Springer}
}

@inproceedings{zambon2020graph,
  title={Graph random neural features for distance-preserving graph representations},
  author={Zambon, Daniele and Alippi, Cesare and Livi, Lorenzo},
  booktitle={International Conference on Machine Learning},
  pages={10968--10977},
  year={2020},
  organization={PMLR}
}

@book{albrecht2024multi,
  title={Multi-agent reinforcement learning: Foundations and modern approaches},
  author={Albrecht, Stefano V and Christianos, Filippos and Sch{\"a}fer, Lukas},
  year={2024},
  publisher={MIT Press}
}

@book{oliehoek2016concise,
  title={A concise introduction to decentralized POMDPs},
  author={Oliehoek, Frans A and Amato, Christopher and others},
  volume={1},
  year={2016},
  publisher={Springer}
}

@incollection{littman1994markov,
  title={Markov games as a framework for multi-agent reinforcement learning},
  author={Littman, Michael L},
  booktitle={Machine learning proceedings 1994},
  pages={157--163},
  year={1994},
  publisher={Elsevier}
}

@inproceedings{gammelli2021graph,
  title={Graph neural network reinforcement learning for autonomous mobility-on-demand systems},
  author={Gammelli, Daniele and Yang, Kaidi and Harrison, James and Rodrigues, Filipe and Pereira, Francisco C and Pavone, Marco},
  booktitle={2021 60th IEEE Conference on Decision and Control (CDC)},
  pages={2996--3003},
  year={2021},
  organization={IEEE}
}

@article{hassouna2026graph,
  title={Graph reinforcement learning for power grids: A comprehensive survey},
  author={Hassouna, Mohamed and Holzh{\"u}ter, Clara and Lytaev, Pawel and Thomas, Josephine and Sick, Bernhard and Scholz, Christoph},
  journal={Energy and AI},
  pages={100671},
  year={2026},
  publisher={Elsevier}
}

@inproceedings{duan2025bayesian,
title={Bayesian Ego-graph inference for Networked Multi-Agent Reinforcement Learning},
author={Wei Duan and Jie Lu and Junyu Xuan},
booktitle={The Thirty-ninth Annual Conference on Neural Information Processing Systems},
year={2025},
url={https://openreview.net/forum?id=3qeTs05bRL}
}

@inproceedings{wang2018nervenet,
  title={Nervenet: Learning structured policy with graph neural networks},
  author={Wang, Tingwu and Liao, Renjie and Ba, Jimmy and Fidler, Sanja},
  booktitle={International conference on learning representations},
  year={2018}
}

@article{marzi2024feudal,
title={Feudal Graph Reinforcement Learning},
author={Tommaso Marzi and Arshjot Singh Khehra and Andrea Cini and Cesare Alippi},
journal={Transactions on Machine Learning Research},
issn={2835-8856},
year={2024},
url={https://openreview.net/forum?id=wFcyJTik90},
note={}
}

@inproceedings{zhang2018fully,
  title={Fully decentralized multi-agent reinforcement learning with networked agents},
  author={Zhang, Kaiqing and Yang, Zhuoran and Liu, Han and Zhang, Tong and Basar, Tamer},
  booktitle={International conference on machine learning},
  pages={5872--5881},
  year={2018},
  organization={PMLR}
}

@inproceedings{
mesbahi2025position,
title={Position: Lifetime tuning is incompatible with continual reinforcement learning},
author={Golnaz Mesbahi and Parham Mohammad Panahi and Olya Mastikhina and Steven Tang and Martha White and Adam White},
booktitle={Forty-second International Conference on Machine Learning Position Paper Track},
year={2025},
url={https://openreview.net/forum?id=JMoWFkwnvv}
}

@software{JaxInforMARL,
      title={JaxInforMARL: Multi-Agent Target MPE RL Environments with GNNs in JAX},
      author={Joseph Selvaraaj},
      year = {2025},
      url = {https://github.com/jselvaraaj/JaxInforMARL},
      version = {1.0.0}
    }

@inproceedings{zheng2024selective,
  title={Selective experience replay compression using coresets for lifelong deep reinforcement learning in medical imaging},
  author={Zheng, Guangyao and Zhou, Samson and Braverman, Vladimir and Jacobs, Michael A and Parekh, Vishwa Sanjay},
  booktitle={Medical Imaging with Deep Learning},
  pages={1751--1764},
  year={2024},
  organization={PMLR}
}

@InProceedings{powers23a,
  title = 	 {Evaluating Continual Learning on a Home Robot},
  author =       {Powers, Sam and Gupta, Abhinav and Paxton, Chris},
  booktitle = 	 {Proceedings of The 2nd Conference on Lifelong Learning Agents},
  pages = 	 {493--512},
  year = 	 {2023},
  editor = 	 {Chandar, Sarath and Pascanu, Razvan and Sedghi, Hanie and Precup, Doina},
  volume = 	 {232},
  series = 	 {Proceedings of Machine Learning Research},
  month = 	 {22--25 Aug},
  publisher =    {PMLR},
  url = 	 {https://proceedings.mlr.press/v232/powers23a.html}
}

@article{chen2026multi,
  title={Multi-Agent Deep Reinforcement Learning (MADRL)-Based End-to-End Formation Control for UAV Swarm with Dynamic Topology},
  author={Chen, Yanping and Xu, Qingyang and Zhang, Chi and Li, Zhengmao},
  journal={Applied Sciences},
  volume={16},
  number={13},
  pages={6554},
  year={2026},
  publisher={MDPI}
}

@inproceedings{caccia2022new,
title={New Insights on Reducing Abrupt Representation Change in Online Continual Learning},
author={Lucas Caccia and Rahaf Aljundi and Nader Asadi and Tinne Tuytelaars and Joelle Pineau and Eugene Belilovsky},
booktitle={International Conference on Learning Representations},
year={2022},
url={https://openreview.net/forum?id=N8MaByOzUfb}
}

@article{sokar2025continual,
  title={Continual learning in vision-language models via aligned model merging},
  author={Sokar, Ghada and Dziugaite, Gintare Karolina and Arnab, Anurag and Iscen, Ahmet and Castro, Pablo Samuel and Schmid, Cordelia},
  journal={arXiv preprint arXiv:2506.03189},
  year={2025}
}

@inproceedings{
wolczyk2021continual,
title={Continual World: A Robotic Benchmark For Continual Reinforcement Learning},
author={Maciej Wolczyk and Micha{\l} Zaj{\k{a}}c and Razvan Pascanu and {\L}ukasz Kuci{\'n}ski and Piotr Mi{\l}o{\'s}},
booktitle={Advances in Neural Information Processing Systems},
editor={A. Beygelzimer and Y. Dauphin and P. Liang and J. Wortman Vaughan},
year={2021},
url={https://openreview.net/forum?id=5qsptDcsdEj}
}

@article{EWC,
author = {James Kirkpatrick  and Razvan Pascanu  and Neil Rabinowitz  and Joel Veness  and Guillaume Desjardins  and Andrei A. Rusu  and Kieran Milan  and John Quan  and Tiago Ramalho  and Agnieszka Grabska-Barwinska  and Demis Hassabis  and Claudia Clopath  and Dharshan Kumaran  and Raia Hadsell },
title = {Overcoming catastrophic forgetting in neural networks},
journal = {Proceedings of the National Academy of Sciences},
volume = {114},
number = {13},
pages = {3521-3526},
year = {2017},
doi = {10.1073/pnas.1611835114},
URL = {https://www.pnas.org/doi/abs/10.1073/pnas.1611835114},
eprint = {https://www.pnas.org/doi/pdf/10.1073/pnas.1611835114}
}

@article{parisi2019continual,
  title={Continual lifelong learning with neural networks: A review},
  author={Parisi, German I and Kemker, Ronald and Part, Jose L and Kanan, Christopher and Wermter, Stefan},
  journal={Neural networks},
  volume={113},
  pages={54--71},
  year={2019},
  publisher={Elsevier}
}

@inproceedings{bettini2023heterogeneous,
  title={Heterogeneous Multi-Robot Reinforcement Learning},
  author={Bettini, Matteo and Shankar, Ajay and Prorok, Amanda},
  booktitle={Proceedings of the 2023 International Conference on Autonomous Agents and Multiagent Systems},
  pages={1485--1494},
  year={2023}
}

@inproceedings{abel2018policy,
  title={Policy and value transfer in lifelong reinforcement learning},
  author={Abel, David and Jinnai, Yuu and Guo, Sophie Yue and Konidaris, George and Littman, Michael},
  booktitle={International conference on machine learning},
  pages={20--29},
  year={2018},
  organization={PMLR}
}

@article{khetarpal2022towards,
  title={Towards continual reinforcement learning: A review and perspectives},
  author={Khetarpal, Khimya and Riemer, Matthew and Rish, Irina and Precup, Doina},
  journal={Journal of Artificial Intelligence Research},
  volume={75},
  pages={1401--1476},
  year={2022}
}

@inproceedings{niu2021multi,
  title={Multi-Agent Graph-Attention Communication and Teaming.},
  author={Niu, Yaru and Paleja, Rohan R and Gombolay, Matthew C},
  booktitle={AAMAS},
  volume={21},
  pages={20th},
  year={2021}
}

@inproceedings{zhou2021overcoming,
  title={Overcoming catastrophic forgetting in graph neural networks with experience replay},
  author={Zhou, Fan and Cao, Chengtai},
  booktitle={Proceedings of the AAAI conference on artificial intelligence},
  volume={35},
  number={5},
  pages={4714--4722},
  year={2021}
}

@article{shi2020masked,
  title={Masked label prediction: Unified message passing model for semi-supervised classification},
  author={Shi, Yunsheng and Huang, Zhengjie and Feng, Shikun and Zhong, Hui and Wang, Wenjin and Sun, Yu},
  journal={arXiv preprint arXiv:2009.03509},
  year={2020}
}

@inproceedings{schneider1999distributed,
  title={Distributed Value Functions},
  author={Schneider, Jeff G and Wong, Weng-Keen and Moore, Andrew W and Riedmiller, Martin A},
  booktitle={Proceedings of the Sixteenth International Conference on Machine Learning},
  pages={371--378},
  year={1999}
}

@inproceedings{Jiang2020Graph,
title={Graph Convolutional Reinforcement Learning},
author={Jiechuan Jiang and Chen Dun and Tiejun Huang and Zongqing Lu},
booktitle={International Conference on Learning Representations},
year={2020},
url={https://openreview.net/forum?id=HkxdQkSYDB}
}

@inproceedings{hu2024learning,
title={Learning Multi-Agent Communication from Graph Modeling Perspective},
author={Shengchao Hu and Li Shen and Ya Zhang and Dacheng Tao},
booktitle={The Twelfth International Conference on Learning Representations},
year={2024},
url={https://openreview.net/forum?id=Qox9rO0kN0}
}

@article{jiang2018learning,
  title={Learning attentional communication for multi-agent cooperation},
  author={Jiang, Jiechuan and Lu, Zongqing},
  journal={Advances in neural information processing systems},
  volume={31},
  year={2018}
}

@article{sukhbaatar2016learning,
  title={Learning multiagent communication with backpropagation},
  author={Sukhbaatar, Sainbayar and Fergus, Rob and others},
  journal={Advances in neural information processing systems},
  volume={29},
  year={2016}
}

@article{foerster2016learning,
  title={Learning to communicate with deep multi-agent reinforcement learning},
  author={Foerster, Jakob and Assael, Ioannis Alexandros and De Freitas, Nando and Whiteson, Shimon},
  journal={Advances in neural information processing systems},
  volume={29},
  year={2016}
}

@inproceedings{das2019tarmac,
  title={Tarmac: Targeted multi-agent communication},
  author={Das, Abhishek and Gervet, Th{\'e}ophile and Romoff, Joshua and Batra, Dhruv and Parikh, Devi and Rabbat, Mike and Pineau, Joelle},
  booktitle={International Conference on machine learning},
  pages={1538--1546},
  year={2019},
  organization={PMLR}
}

@inproceedings{agarwal2020learning,
author = {Agarwal, Akshat and Kumar, Sumit and Sycara, Katia and Lewis, Michael},
title = {Learning Transferable Cooperative Behavior in Multi-Agent Teams},
year = {2020},
isbn = {9781450375184},
publisher = {International Foundation for Autonomous Agents and Multiagent Systems},
address = {Richland, SC},
booktitle = {Proceedings of the 19th International Conference on Autonomous Agents and MultiAgent Systems},
pages = {1741–1743},
numpages = {3},
location = {Auckland, New Zealand},
series = {AAMAS '20}
}

@inproceedings{chaudhry2018efficient,
title={Efficient Lifelong Learning with A-{GEM}},
author={Arslan Chaudhry and Marc’Aurelio Ranzato and Marcus Rohrbach and Mohamed Elhoseiny},
booktitle={International Conference on Learning Representations},
year={2019},
url={https://openreview.net/forum?id=Hkf2_sC5FX},
}

@inproceedings{MAS,
author = {Aljundi, Rahaf and Babiloni, Francesca and Elhoseiny, Mohamed and Rohrbach, Marcus and Tuytelaars, Tinne},
title = {Memory Aware Synapses: Learning What (not) to Forget},
year = {2018},
isbn = {978-3-030-01218-2},
publisher = {Springer-Verlag},
address = {Berlin, Heidelberg},
url = {https://doi.org/10.1007/978-3-030-01219-9_9},
doi = {10.1007/978-3-030-01219-9_9},
booktitle = {Computer Vision – ECCV 2018: 15th European Conference, Munich, Germany, September 8–14, 2018, Proceedings, Part III},
pages = {144–161},
numpages = {18},
location = {Munich, Germany}
}

@article{bacciu2020gentle,
  title={A gentle introduction to deep learning for graphs},
  author={Bacciu, Davide and Errica, Federico and Micheli, Alessio and Podda, Marco},
  journal={Neural Networks},
  volume={129},
  pages={203--221},
  year={2020},
  publisher={Elsevier}
}

@inproceedings{gilmer2017neural,
  title={Neural message passing for quantum chemistry},
  author={Gilmer, Justin and Schoenholz, Samuel S and Riley, Patrick F and Vinyals, Oriol and Dahl, George E},
  booktitle={International conference on machine learning},
  pages={1263--1272},
  year={2017},
  organization={Pmlr}
}

@article{zhu2024survey,
  title={A survey of multi-agent deep reinforcement learning with communication},
  author={Zhu, Changxi and Dastani, Mehdi and Wang, Shihan},
  journal={Autonomous Agents and Multi-Agent Systems},
  volume={38},
  number={1},
  pages={4},
  year={2024},
  publisher={Springer}
}

@article{yuan2024multiagent,
  title={Multiagent continual coordination via progressive task contextualization},
  author={Yuan, Lei and Li, Lihe and Zhang, Ziqian and Zhang, Fuxiang and Guan, Cong and Yu, Yang},
  journal={IEEE Transactions on Neural Networks and Learning Systems},
  volume={36},
  number={4},
  pages={6326--6340},
  year={2024},
  publisher={IEEE}
}

@article{qu2019value,
  title={Value propagation for decentralized networked deep multi-agent reinforcement learning},
  author={Qu, Chao and Mannor, Shie and Xu, Huan and Qi, Yuan and Song, Le and Xiong, Junwu},
  journal={Advances in Neural Information Processing Systems},
  volume={32},
  year={2019}
}

@article{yi2022learning,
  title={Learning to share in networked multi-agent reinforcement learning},
  author={Yi, Yuxuan and Li, Ge and Wang, Yaowei and Lu, Zongqing},
  journal={Advances in Neural Information Processing Systems},
  volume={35},
  pages={15119--15131},
  year={2022}
}

@article{qu2022scalable,
  title={Scalable reinforcement learning for multiagent networked systems},
  author={Qu, Guannan and Wierman, Adam and Li, Na},
  journal={Operations Research},
  volume={70},
  number={6},
  pages={3601--3628},
  year={2022},
  publisher={INFORMS}
}

@article{lin2021multi,
  title={Multi-agent reinforcement learning in stochastic networked systems},
  author={Lin, Yiheng and Qu, Guannan and Huang, Longbo and Wierman, Adam},
  journal={Advances in neural information processing systems},
  volume={34},
  pages={7825--7837},
  year={2021}
}

@inproceedings{rashwan2026factored,
title={Factored Value Functions for Graph-Based Multi-Agent Reinforcement Learning},
author={Ahmed Rashwan and Keith Briggs and Chris Budd and Lisa Maria Kreusser},
booktitle={Forty-third International Conference on Machine Learning},
year={2026},
url={https://openreview.net/forum?id=gNwwzeqPF0}
}

@article{liu2025survey,
  title={Survey on graph-based reinforcement learning for networked coordination and control},
  author={Liu, Yifan and Wu, Dalei and Liang, Yu},
  journal={Automation},
  volume={6},
  number={4},
  pages={65},
  year={2025},
  publisher={MDPI}
}

@inproceedings{singh2018individualized,
title={Learning when to Communicate at Scale in Multiagent Cooperative and Competitive Tasks},
author={Amanpreet Singh and Tushar Jain and Sainbayar Sukhbaatar},
booktitle={International Conference on Learning Representations},
year={2019},
url={https://openreview.net/forum?id=rye7knCqK7},
}

@inproceedings{yoon2021winning,
title={Winning the L2{\{}RPN{\}} Challenge: Power Grid Management via Semi-Markov Afterstate Actor-Critic},
author={Deunsol Yoon and Sunghoon Hong and Byung-Jun Lee and Kee-Eung Kim},
booktitle={International Conference on Learning Representations},
year={2021},
url={https://openreview.net/forum?id=LmUJqB1Cz8}
}

@INPROCEEDINGS{8569301,
  author={Nishi, Tomoki and Otaki, Keisuke and Hayakawa, Keiichiro and Yoshimura, Takayoshi},
  booktitle={2018 21st International Conference on Intelligent Transportation Systems (ITSC)}, 
  title={Traffic Signal Control Based on Reinforcement Learning with Graph Convolutional Neural Nets}, 
  year={2018},
  volume={},
  number={},
  pages={877-883},
  doi={10.1109/ITSC.2018.8569301}}

@ARTICLE{9405489,
  author={Devailly, François-Xavier and Larocque, Denis and Charlin, Laurent},
  journal={IEEE Transactions on Intelligent Transportation Systems}, 
  title={IG-RL: Inductive Graph Reinforcement Learning for Massive-Scale Traffic Signal Control}, 
  year={2022},
  volume={23},
  number={7},
  pages={7496-7507},
  doi={10.1109/TITS.2021.3070835}}

@inproceedings{nayak2023scalable,
  title={Scalable multi-agent reinforcement learning through intelligent information aggregation},
  author={Nayak, Siddharth and Choi, Kenneth and Ding, Wenqi and Dolan, Sydney and Gopalakrishnan, Karthik and Balakrishnan, Hamsa},
  booktitle={International conference on machine learning},
  pages={25817--25833},
  year={2023},
  organization={PMLR}
}

@article{marzi2025hierarchical,
  title={Hierarchical Message-Passing Policies for Multi-Agent Reinforcement Learning},
  author={Marzi, Tommaso and Alippi, Cesare and Cini, Andrea},
  journal={arXiv preprint arXiv:2507.23604},
  year={2025}
}

@inproceedings{candela2022transferring,
  title={Transferring multi-agent reinforcement learning policies for autonomous driving using sim-to-real},
  author={Candela, Eduardo and Parada, Leandro and Marques, Luis and Georgescu, Tiberiu-Andrei and Demiris, Yiannis and Angeloudis, Panagiotis},
  booktitle={2022 IEEE/RSJ International Conference on Intelligent Robots and Systems (IROS)},
  pages={8814--8820},
  year={2022},
  organization={IEEE}
}

@article{kolat2023multi,
  title={Multi-agent reinforcement learning for traffic signal control: A cooperative approach},
  author={Kolat, M{\'a}t{\'e} and K{\H{o}}v{\'a}ri, B{\'a}lint and B{\'e}csi, Tam{\'a}s and Aradi, Szil{\'a}rd},
  journal={Sustainability},
  volume={15},
  number={4},
  pages={3479},
  year={2023},
  publisher={MDPI}
}

@article{vinyals2019grandmaster,
  title={Grandmaster level in StarCraft II using multi-agent reinforcement learning},
  author={Vinyals, Oriol and Babuschkin, Igor and Czarnecki, Wojciech M and Mathieu, Micha{\"e}l and Dudzik, Andrew and Chung, Junyoung and Choi, David H and Powell, Richard and Ewalds, Timo and Georgiev, Petko and others},
  journal={nature},
  volume={575},
  number={7782},
  pages={350--354},
  year={2019},
  publisher={Nature Publishing Group UK London}
}

@article{jaeger2004harnessing,
  title={Harnessing nonlinearity: Predicting chaotic systems and saving energy in wireless communication},
  author={Jaeger, Herbert and Haas, Harald},
  journal={science},
  volume={304},
  number={5667},
  pages={78--80},
  year={2004},
  publisher={American Association for the Advancement of Science}
}

@article{pao1992neural,
  title={Neural-net computing and the intelligent control of systems},
  author={Pao, Yoh-Han and Phillips, Stephen M and Sobajic, Dejan J},
  journal={International Journal of Control},
  volume={56},
  number={2},
  pages={263--289},
  year={1992},
  publisher={Taylor \& Francis}
}

@article{rahimi2007random,
  title={Random features for large-scale kernel machines},
  author={Rahimi, Ali and Recht, Benjamin},
  journal={Advances in neural information processing systems},
  volume={20},
  year={2007}
}

@article{li2023crnet,
  title={CRNet: A fast continual learning framework with random theory},
  author={Li, Depeng and Zeng, Zhigang},
  journal={IEEE Transactions on Pattern Analysis and Machine Intelligence},
  volume={45},
  number={9},
  pages={10731--10744},
  year={2023},
  publisher={IEEE}
}

@article{mcdonnell2023ranpac,
  title={Ranpac: Random projections and pre-trained models for continual learning},
  author={McDonnell, Mark D and Gong, Dong and Parvaneh, Amin and Abbasnejad, Ehsan and Van den Hengel, Anton},
  journal={Advances in Neural Information Processing Systems},
  volume={36},
  pages={12022--12053},
  year={2023}
}

@article{prabhu2024randumb,
  title={RanDumb: Random representations outperform online continually learned representations},
  author={Prabhu, Ameya and Sinha, Shiven and Kumaraguru, Ponnurangam and Torr, Philip H and Sener, Ozan and Dokania, Puneet K},
  journal={Advances in Neural Information Processing Systems},
  volume={37},
  pages={37988--38006},
  year={2024}
}

@article{yu2022surprising,
  title={The surprising effectiveness of ppo in cooperative multi-agent games},
  author={Yu, Chao and Velu, Akash and Vinitsky, Eugene and Gao, Jiaxuan and Wang, Yu and Bayen, Alexandre and Wu, Yi},
  journal={Advances in neural information processing systems},
  volume={35},
  pages={24611--24624},
  year={2022}
}

@inproceedings{
burda2018largescale,
title={Large-Scale Study of Curiosity-Driven Learning},
author={Yuri Burda and Harri Edwards and Deepak Pathak and Amos Storkey and Trevor Darrell and Alexei A. Efros},
booktitle={International Conference on Learning Representations},
year={2019},
url={https://openreview.net/forum?id=rJNwDjAqYX},
}

@inproceedings{
burda2018exploration,
title={Exploration by random network distillation},
author={Yuri Burda and Harrison Edwards and Amos Storkey and Oleg Klimov},
booktitle={International Conference on Learning Representations},
year={2019},
url={https://openreview.net/forum?id=H1lJJnR5Ym},
}

@article{osband2018randomized,
  title={Randomized prior functions for deep reinforcement learning},
  author={Osband, Ian and Aslanides, John and Cassirer, Albin},
  journal={Advances in neural information processing systems},
  volume={31},
  year={2018}
}

@article{chen2023self,
  title={Self-supervised reinforcement learning that transfers using random features},
  author={Chen, Boyuan and Zhu, Chuning and Agrawal, Pulkit and Zhang, Kaiqing and Gupta, Abhishek},
  journal={Advances in Neural Information Processing Systems},
  volume={36},
  pages={56411--56436},
  year={2023}
}

@inproceedings{yang2024transfer,
  title={A transfer approach using graph neural networks in deep reinforcement learning},
  author={Yang, Tianpei and You, Heng and Hao, Jianye and Zheng, Yan and Taylor, Matthew E},
  booktitle={Proceedings of the AAAI conference on artificial intelligence},
  volume={38},
  number={15},
  pages={16352--16360},
  year={2024}
}

@incollection{mccloskey1989catastrophic,
  title={Catastrophic interference in connectionist networks: The sequential learning problem},
  author={McCloskey, Michael and Cohen, Neal J},
  booktitle={Psychology of learning and motivation},
  volume={24},
  pages={109--165},
  year={1989},
  publisher={Elsevier}
}

@article{nikishin2023deep,
  title={Deep reinforcement learning with plasticity injection},
  author={Nikishin, Evgenii and Oh, Junhyuk and Ostrovski, Georg and Lyle, Clare and Pascanu, Razvan and Dabney, Will and Barreto, Andr{\'e}},
  journal={Advances in Neural Information Processing Systems},
  volume={36},
  pages={37142--37159},
  year={2023}
}

@inproceedings{hayes2020lifelong,
  title={Lifelong machine learning with deep streaming linear discriminant analysis},
  author={Hayes, Tyler L and Kanan, Christopher},
  booktitle={2020 IEEE/CVF Conference on Computer Vision and Pattern Recognition Workshops (CVPRW)},
  pages={887--896},
  year={2020},
  organization={IEEE}
}

@inproceedings{amato2025encoding,
  title={Encoding goals as graphs: Structured objectives for scalable cooperative multi-agent reinforcement learning},
  author={Amato, Alessandro and Galliera, Raffaele and Venable, K Brent and Suri, Niranjan},
  booktitle={Second Coordination and Cooperation in Multi-Agent Reinforcement Learning Workshop},
  year={2025}
}

@inproceedings{chen2020simple,
  title={A simple framework for contrastive learning of visual representations},
  author={Chen, Ting and Kornblith, Simon and Norouzi, Mohammad and Hinton, Geoffrey},
  booktitle={International conference on machine learning},
  pages={1597--1607},
  year={2020},
  organization={PmLR}
}

@inproceedings{kolesnikov2019revisiting,
  title={Revisiting self-supervised visual representation learning},
  author={Kolesnikov, Alexander and Zhai, Xiaohua and Beyer, Lucas},
  booktitle={2019 IEEE/CVF Conference on Computer Vision and Pattern Recognition (CVPR)},
  pages={1920--1929},
  year={2019},
  organization={IEEE}
}

@inproceedings{houlsby2019parameter,
  title={Parameter-efficient transfer learning for NLP},
  author={Houlsby, Neil and Giurgiu, Andrei and Jastrzebski, Stanislaw and Morrone, Bruna and De Laroussilhe, Quentin and Gesmundo, Andrea and Attariyan, Mona and Gelly, Sylvain},
  booktitle={International conference on machine learning},
  pages={2790--2799},
  year={2019},
  organization={PMLR}
}

@InProceedings{R3M,
  title = 	 {R3M: A Universal Visual Representation for Robot Manipulation},
  author =       {Nair, Suraj and Rajeswaran, Aravind and Kumar, Vikash and Finn, Chelsea and Gupta, Abhinav},
  booktitle = 	 {Proceedings of The 6th Conference on Robot Learning},
  pages = 	 {892--909},
  year = 	 {2023},
  editor = 	 {Liu, Karen and Kulic, Dana and Ichnowski, Jeff},
  volume = 	 {205},
  series = 	 {Proceedings of Machine Learning Research},
  month = 	 {14--18 Dec},
  publisher =    {PMLR},
  url = 	 {https://proceedings.mlr.press/v205/nair23a.html}
}

@inproceedings{
ma2023vip,
title={{VIP}: Towards Universal Visual Reward and Representation via Value-Implicit Pre-Training},
author={Yecheng Jason Ma and Shagun Sodhani and Dinesh Jayaraman and Osbert Bastani and Vikash Kumar and Amy Zhang},
booktitle={The Eleventh International Conference on Learning Representations },
year={2023},
url={https://openreview.net/forum?id=YJ7o2wetJ2}
}

@article{rusu2016progressive,
  title={Progressive neural networks},
  author={Rusu, Andrei A and Rabinowitz, Neil C and Desjardins, Guillaume and Soyer, Hubert and Kirkpatrick, James and Kavukcuoglu, Koray and Pascanu, Razvan and Hadsell, Raia},
  journal={arXiv preprint arXiv:1606.04671},
  year={2016}
}

@inproceedings{powers2022self,
  title={Self-activating neural ensembles for continual reinforcement learning},
  author={Powers, Sam and Xing, Eliot and Gupta, Abhinav},
  booktitle={Conference on Lifelong Learning Agents},
  pages={683--704},
  year={2022},
  organization={PMLR}
}

@article{zheng2025mastering,
  title={Mastering continual reinforcement learning through fine-grained sparse network allocation and dormant neuron exploration},
  author={Zheng, Chengqi and Yin, Haiyan and Chen, Jianda and Ng, Terence and Ong, Yew-Soon and Tsang, Ivor},
  journal={arXiv preprint arXiv:2503.05246},
  year={2025}
}

@article{hu2026continual,
  title={Continual knowledge adaptation for reinforcement learning},
  author={Hu, Jinwu and Lian, Zihao and Wen, Zhiquan and Li, Chenghao and Chen, Guohao and Wen, Xutao and Xiao, Bin and Tan, Mingkui},
  journal={Advances in Neural Information Processing Systems},
  volume={38},
  pages={36283--36313},
  year={2026}
}

@article{hu2024decentralized,
  title={Decentralized graph-based multi-agent reinforcement learning using reward machines},
  author={Hu, Jueming and Xu, Zhe and Wang, Weichang and Qu, Guannan and Pang, Yutian and Liu, Yongming},
  journal={Neurocomputing},
  volume={564},
  pages={126974},
  year={2024},
  publisher={Elsevier}
}
\bibliographystyle{plainnat}

\clearpage

\appendix
\section{Theoretical Analysis of Forgetting}~\label{app:theory_forgetting}
In this section, we provide a theoretical analysis of how forgetting emerges in a decoupled graph-based policy as in~\autoref{eq:grf_policy}.
For simplicity, we remove the time index $t$ within a task and call $\va \doteq \boldsymbol{\pi}_k(\vo, \gG_h)$ the joint action taken by the joint policy $\boldsymbol{\pi}_k$ evaluated on task $h$ (characterized by graph $\gG_h$) after being trained on task $k$. Therefore, we have:
\begin{equation}
    \va \doteq \boldsymbol{\pi}_k(\vo, \gG_h) = \{\pi_{\theta_k}(\vo^i;\gG_h)\}_{i=1}^\gN
\end{equation}
Given task indexes $k$ and $h$, with $k>h$, forgetting on task $h$ after training on task $k$ is defined as the performance difference:
\begin{align}
    J_h(\boldsymbol{\pi}_k) - J_h(\boldsymbol{\pi}_h) \leq |J_h(\boldsymbol{\pi}_k) - J_h(\boldsymbol{\pi}_h)| \leq K^{\boldsymbol{\pi}} \lVert \boldsymbol{\pi}_k(\vo;\gG_h) - \boldsymbol{\pi}_h(\vo;\gG_h) \rVert_\infty
\end{align}
where we assumed a Lipschitz continuous objective and denoted as $K^{\boldsymbol{\pi}}$ the Lipschitz constant. If we substitute a generic decoupled policy $\boldsymbol{\pi} = \boldsymbol{\mu} \circ \boldsymbol{\psi}$ and use the triangle inequality (sup-norm sub-additivity), we get:
\begin{align}
   &J_h(\boldsymbol{\pi}_k) - J_h(\boldsymbol{\pi}_h) \leq  K^{\boldsymbol{\pi}} \lVert \boldsymbol{\mu}_k(\boldsymbol{\psi}_k(\vo;\gG_h)) - \boldsymbol{\mu}_h(\boldsymbol{\psi}_h(\vo;\gG_h)) \rVert_\infty \nonumber\\
   &\leq\lVert\boldsymbol{\mu}_k(\boldsymbol{\psi}_k(\vo;\gG_h)) - \boldsymbol{\mu}_k(\boldsymbol{\psi}_h(\vo;\gG_h))\rVert_\infty +\lVert\boldsymbol{\mu}_k(\boldsymbol{\psi}_h(\vo;\gG_h)) - \boldsymbol{\mu}_h(\boldsymbol{\psi}_h(\vo;\gG_h))\rVert_\infty 
\end{align}
In the last inequality, the two terms represent the performance drop caused by representation learning and action prediction, respectively. 

\section{Extended Related Work}~\label{app:ext_rw}
\paragraph{Graph-based MARL and Communication}
Graph-based MARL formalizes settings where the environment dynamics depend on the underlying graph. This has been adopted across several settings, e.g., where the dynamics of each agent depend on its neighbors in the graph ~\citep{hu2024decentralized,rashwan2026factored}. The same graph possibly serves as the agent communication structure. Such problems have been extensively studied in networked systems~\citep{zhang2018fully}. For example, NeurComm~\citep{Chu2020Multi-agent} employs an effective communication protocol that prevents information loss during message aggregation. Conversely, BayesG~\citep{duan2025bayesian} learns a context-aware interaction graph that reduces unnecessary message propagation. Other works address networked problems by either leveraging consensus-based mechanisms~\citep{qu2019value,yi2022learning} or incorporating relational information~\citep{lin2021multi, qu2022scalable}. 
Communication MARL~\citep{zhu2024survey} focuses on developing effective methods for improving agents' coordination through communication. Pioneering methods~\citep{foerster2016learning,sukhbaatar2016learning} focused on learning communication protocols end-to-end. Subsequent works~\citep{jiang2018learning,singh2018individualized,das2019tarmac,hu2024learning} employed message-broadcasting schemes that adaptively select communication channels. Other approaches employ GNNs in the decision-making process to learn message-passing policies~\citep{Jiang2020Graph,niu2021multi,bettini2023heterogeneous, marzi2025hierarchical}, possibly focusing on the zero-shot capabilities of the learned behaviors~\citep{agarwal2020learning,nayak2023scalable}. Here, the communication topology is either given by the environment or learned end-to-end, and does not necessarily determine the environment dynamics.

\paragraph{Continual RL} In Continual RL~(CRL;~\citep{khetarpal2022towards,abel2023definition}), the agent learns from a sequence of RL problems, aiming to effectively adapt to new tasks while retaining acquired knowledge from previously observed ones. CRL has been applied to different domains, including robotics~\citep{powers23a} and medical imaging~\citep{zheng2024selective}. A common approach to address CRL problems is to leverage standard CL methods~\citep{pan2025survey}, such as replay-based mechanisms (e.g., AGEM;~\citep{chaudhry2018efficient}), parameter regularizations such as EWC~\citep{EWC} or MAS~\citep{MAS}, or parameter isolation (e.g., PackNet;~\citep{mallya2018packnet}). The Continual MARL~(CMARL;~\citep{tomilin2026meal}) framework extends the CRL problem to systems of multiple cooperative agents, requiring novel methodologies tailored for this paradigm~\citep{yuan2024multiagent}.

\paragraph{Frozen Representations} 
Frozen representations have proved their effectiveness as both purely random features~\citep{pao1992neural,jaeger2004harnessing,rahimi2007random} and as pre-trained features~\citep{houlsby2019parameter,kolesnikov2019revisiting,chen2020simple}. In Reinforcement Learning~(RL), random features have been used to improve exploration~\citep{burda2018largescale,burda2018exploration}, maintain accurate uncertainty estimates~\citep{osband2018randomized}, or generate intrinsic learning signals~\citep{chen2023self}. In contrast, pre-trained representations have been successfully applied to different domains, such as visual RL~\citep{R3M,ma2023vip} or graph-based MARL~\citep{amato2025encoding}.
Frozen representations have also been explored in Continual Learning problems~\citep{hayes2020lifelong,li2023crnet,mcdonnell2023ranpac, prabhu2024randumb}. Notably,~\citet{donghi2025unreasonable} leverage randomized representations generated by different GNN backbones~\citep{zambon2020graph,navarin2023untrained} in CL problems on graphs~\citep{donghi2025online}. In Continual RL, several works rely on frozen representations or modules~\citep{rusu2016progressive,powers2022self,nikishin2023deep,hu2026continual} to mitigate forgetting and facilitate learning by reducing the number of learnable parameters~\citep{zheng2025mastering}. Nonetheless, to the best of our knowledge, no existing approaches have investigated frozen GNN-based encoders for graph-based CMARL settings. 

\section{\underline{Gra}ph-based \underline{Fo}rmation (GRAFO) Benchmark}~\label{app:cgmarl_environment}
In this section, we present our novel graph-based MARL benchmark, named \GRAFO. 
Building on other formation environments~\citep{agarwal2020learning}, GRAFO is a collection of graph-based MARL tasks in which the agents' objective is to navigate to a target configuration that resembles the input graph. Notably, such configurations can be generated arbitrarily, making GRAFO a suitable benchmark for CGMARL problems. 

\subsection{Single-Task Description}
\begin{figure*}[t]
\includegraphics[width=1.\textwidth]{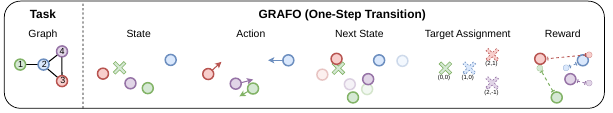}
\vspace{-.6cm}
    \caption{One-step transition in GRAFO. At the beginning of the episode, an initial state in which the agents (colored circles) and known landmark (solid, green cross) positions are randomly assigned is generated. Then, at each step, agents take actions depending on the current state. Each agent's reward is the negative of the distance to its corresponding target (dashed cross with the same color). In the case of two symmetric agents, i.e., with a non-empty set $\gK$ of structurally-equivalent nodes, a linear sum assignment problem is solved to assign the targets. }
\label{fig:magma_transition}
\end{figure*}

In GRAFO, the goal of the agents is to navigate to a \textit{partially unknown} target configuration, which depends on the input graph. 
A single task configuration is specified by:
\begin{itemize}
    \item the adjacency matrix $A$ of the graph, which influences the environment dynamics and possibly serves as the communication structure (\autoref{sec:preliminaries}). 
    \item a tuple of target distances $\vg = (\vg_1,\dots,\vg_N)$, indicating the relative position shifts with respect to the known landmark in the $2$-dimensional plane; a target distance $\vg_i = (0,0)$ corresponds to the reference node, i.e., the node assigned to the known landmark. 
    \item possibly, a set of node indices $\gK\subset \gN$ that are structurally equivalent in the graph.
    \item optionally, static graph (node and edge) features $\mX$ and $\mE$ (please refer to~\autoref{sec:single_task}).
\end{itemize}
We report an example for a one-step transition with a $4$-agent task in~\autoref{fig:magma_transition}. The corresponding configuration is the following:
\begin{equation}~\label{eq:grafo_single}
    A = \begin{pmatrix}
0 & 1 & 0 & 0\\
1 & 0 & 1 & 1\\
0 & 1 & 0 & 1\\
0 & 1 & 1 & 0\\
\end{pmatrix}
\qquad
\vg = \Bigl((0,0), (1, 0), (2, 1), (2,-1)\Bigr)
\qquad
\gK = \{3,4\} \subset \gN = \{1,2,3,4\}
\end{equation}

Similar to the Formation environment~\citep{agarwal2020learning}, all agents have access to the position of a reference landmark (green cross), which serves as the target position for a specific agent (green circle). In the example, agent $1$ (green) is the reference node, as its goal shift is $\vg_1 = (0,0)$, meaning that its target position coincides with the known landmark. Notably, the reference agent is \textit{unknown} to the team. Importantly, in GRAFO, the target positions (dashed crosses) of other agents must be deduced from the underlying graph (adjacency matrix $A$, and optionally, static graph features), rather than being explicitly provided. In other words, while agents in Formation must arrange themselves in a symmetric configuration around the landmark, in GRAFO, the graph structure is crucial for determining the target positions $\vg$ and solving the task. We assess this claim in~\autoref{app:single_task_performance}.

Each agent's reward is the negative distance from its landmark, plus a one-time bonus whenever an agent reaches its target position. When two agents are structurally equivalent (i.e., symmetric under a graph automorphism), their target positions are assigned at each step to candidate landmarks using a linear sum assignment problem that minimizes the total distance. In the example, agents $3$ and $4$ can be swapped without altering the graph topology, and therefore, they belong to the set $\gK$ of structurally equivalent nodes. This implies that at each step, target positions $\vg_3 = (2, 1)$ and $\vg_4 = (2,-1)$ are assigned to agent $3$ and $4$ depending on the configuration that minimizes the total distance. Lastly, the position of the known landmark is randomly initialized at the beginning of each episode, and other target positions are assigned using $\vg$ accordingly.

\paragraph{Graph-based MARL Formulation}
We note that the adjacency $A$ and the target positions $\vg$ determine the reward function $\gR_\gG$, which makes GRAFO a proper graph-based MARL problem~(\autoref{eq:nmarl}). Indeed, each agent's reward is defined based on the distance to its target position, which changes depending on $A$ and $\vg$. Consequently, standard GNN-based policies are likely to fail when applied in a zero-shot manner, and we assess this behavior in~\autoref{sec:forgetting_cgmarl}. The transition function, as well as the observation and action spaces, are identical to those of the Formation environment from~\citet{agarwal2020learning}.

\subsection{Continual GRAFO}
In line with Definition~\ref{def:cgmarl}, we can define a task sequence by simply considering a series of different graphs. This implies that we can procedurally generate (possibly infinite) CGMARL sequences by simply varying the task specification in~\autoref{eq:grafo_single}. We now show how the different CGMARL settings (\autoref{sec:cgmarl}) can be induced in GRAFO. The complete sequences used in our experiments are reported in~\autoref{sec:task_sequences}.

\textbf{Topology Switch} In Topology Switch~(\autoref{sec:ts_scenario}), different tasks in the TS sequence are specified by different topologies, i.e., nodes and/or edges. For example, a $2$-task TS sequence can be defined as:
\begin{align*}
     A_1 &= \begin{pmatrix}
0 & 1 & 0 & 0\\
1 & 0 & 1 & 1\\
0 & 1 & 0 & 1\\
0 & 1 & 1 & 0\\
\end{pmatrix}
\qquad
\vg_1 = \Bigl((0,0), (1, 0), (2, 0.5), (2,-0.5)\Bigr)
\qquad
\gK_1 = \{3,4\}\\
 A_2 &= \begin{pmatrix}
0 & 1 & 0 \\
1 & 0 & 1 \\
0 & 1 & 0 \\
\end{pmatrix}
\qquad
\vg_2 = \Bigl((-1,0),(0,0), (1, 0)\Bigr)
\qquad
\gK_2 = \{1,3\}
\end{align*}
where the first task has $4$ agents, while the second one has $3$ (please refer to GRAFO-$4(1)$ and GRAFO-$3$ in~\autoref{fig:grafo_tasks} for a visual representation).

\textbf{Connectivity Switch} In Connectivity Switch~(\autoref{sec:cs_scenario}), different tasks in the CS sequence are specified by different edges, while the node set is fixed. For example, a $2$-task CS sequence can be defined as:
\begin{align*}
     A_1 &= \begin{pmatrix}
0 & 1 & 0 & 0 & 0 & 0\\
1 & 0 & 1 & 0 & 0& 0\\
0 & 1 & 0 & 1 & 0& 0 \\
0 & 0 & 1 & 0 & 1& 1\\
0 & 0 & 0 & 1 & 0& 1\\
0 & 0 & 0 & 1 & 1& 0\\
\end{pmatrix}
\quad
\vg_1 = \Bigl(
    (0,0),
    (1, 0),
    (2, 0),
    (3, 0),
    (4, 0.5),
     (4,-0.5)\Bigr)
\quad
\gK_1 = \{5,6\}\\
 A_2 &= \begin{pmatrix}
0 & 1 & 0 & 0 & 0 & 0 \\
1 & 0 & 1 & 0 & 0 & 0 \\
0 & 1 & 0 & 1 & 1 & 0\\
0 & 0 & 1 & 0 & 0 & 1 \\
0 & 0 & 1 & 0 & 0 & 1\\
0 & 0 & 0 & 1 & 1 & 0\\
\end{pmatrix}
\quad
\vg_2 = \Bigl(
    (0,0),
    (1, 0),
    (2, 0),
    (3, 0.5),
    (4, 0),
     (4,0.5)\Bigr)
\quad
\gK_2 = \{4,5\}
\end{align*}
where, in this case, the sequence is specified by $6$ agents for both tasks (please refer to GRAFO-$6(1)$ and GRAFO-$6(2)$ in~\autoref{fig:grafo_tasks} for a visual representation).

\textbf{Graph Features Switch} In Graph Features Switch~(\autoref{sec:gfs_scenario}), different tasks in the GFS sequence are specified by different graph features. For example, a $2$-task GFS sequence can be defined as:
\begin{align*}
     A_1 &= \begin{pmatrix}
0 & 1 & 0 & 0\\
1 & 0 & 1 & 1\\
0 & 1 & 0 & 1\\
0 & 1 & 1 & 0\\
\end{pmatrix}
\qquad
\vg_1 = \Bigl((0,0), (1, 0), (2, 0.5), (2,-0.5)\Bigr)
\qquad
\gK_1 = \{3,4\}\\
A_2 &= \begin{pmatrix}
0 & 1 & 0 & 0\\
1 & 0 & 1 & 1\\
0 & 1 & 0 & 1\\
0 & 1 & 1 & 0\\
\end{pmatrix}
\qquad
\vg_2 = \Bigl((0,0), (1, 0), (2, 1.5), (2,-1.5)\Bigr)
\qquad
\gK_2 = \{3,4\}\\
\end{align*}
where, in this case, the sequence is characterized by $4$ agents with fixed edges (please refer to GRAFO-$4(1)$ and GRAFO-$4(2)$ in~\autoref{fig:grafo_tasks} for a visual representation).
We note that, in the GFS scenario, the topology is fixed across the sequence, and the graph features are used to distinguish different tasks. Therefore, in GRAFO, the target positions $\vg$ must be encoded within the policy. This can be achieved by, e.g., providing the relative position shift as node features. In the example above, the first task would have node feature $1$, while the second one would have node feature $2$. Another option is to provide the target $L^2$-distances as edge features.

\section{Additional Results}

\subsection{Single-Task Performance in GRAFO}~\label{app:single_task_performance}
In~\autoref{tab:TS_single_task_resolution}, we report the performance achieved by InforMARL~(\autoref{app:informarl}) on each of the single tasks of the TS sequence (\autoref{sec:task_sequences}). In particular, we consider InforMARL and compare it with a variant where the policy does not rely on a GNN, i.e., without message exchange across agents. The results show that the graph-based version can correctly solve each task individually, while the variant without GNN fails to do so. This supports our claim that GNN-based policies are crucial for effectively solving individual graph-based tasks in the GRAFO benchmark. 
\def\thickhline{\noalign{\hrule height1.3pt}}

\begin{table}[h]
\vspace{-0.3cm}
\caption{Solved percentage of InforMARL and the variant without GNN in the actor on the single tasks of the TS sequence (reported as median $\pm$ standard deviation of $5$ seeds). }
\vspace{0.2cm}
\label{tab:TS_single_task_resolution}
\centering
\resizebox{1.0\columnwidth}{!}{
\begin{tabular}{c|c|c|c|c|c|c}
\thickhline
 & GRAFO-$3$ & GRAFO-$4$& GRAFO-$5$ & GRAFO-$6$ (1) & GRAFO-$6$ (3) & GRAFO-$7$  \\
\thickhline
InforMARL &  $\boldsymbol{1}\pm \scriptstyle{0}$ & $\boldsymbol{1}\pm \scriptstyle{0}$& $\boldsymbol{1}\pm \scriptstyle{0}$&$\boldsymbol{0.994}\pm \scriptstyle{0.009}$& $\boldsymbol{1}\pm \scriptstyle{0}$&$\boldsymbol{0.997}\pm \scriptstyle{0.007}$\\
InforMARL (no GNN)&  ${0.531}\pm \scriptstyle{0.013}$ & ${0.328}\pm \scriptstyle{0.091}$& ${0.297}\pm \scriptstyle{0.258}$&${0}\pm \scriptstyle{0}$& ${0.172}\pm \scriptstyle{0.169}$&${0}\pm \scriptstyle{0}$\\
\hline
\end{tabular}
}
\end{table}

\subsection{Impact of Regularization Coefficients}~\label{app:analysis_reg_coeff}
To investigate the results achieved by EWC (M) in~\autoref{fig:full_benchmark}, we perform an additional experiment on the TS sequence to assess the impact of using two separate regularization coefficients. Specifically, we set the coefficient for the graph encoder $\lambda_\psi = 10^{10}$ (as in~\autoref{fig:full_benchmark}), which is the best value achieved during parameter tuning (refer to~\autoref{app:hyperparameters}). In contrast, we vary the coefficient $\lambda_{\mu}$ for the action prediction module from $10^{4}$ to $10^{12}$, and evaluate each configuration $(\lambda_{\psi},\lambda_{\mu})$ on the TS sequence. We note that, when $\lambda_{\mu}  = 10^{10}= \lambda_{\psi}$  (pink box), the results are those reported in~\autoref{fig:full_benchmark}. Our hypothesis to explain the low plasticity of EWC (M) in~\autoref{fig:full_benchmark} is that the graph encoder requires a much stronger regularization coefficient than the action prediction module, since EWC is not designed to preserve structural knowledge. Therefore, despite being a single value in practice, these coefficients lie on two different scales: large coefficients for the action module (large $\lambda_{\mu}$) can deteriorate learning new tasks, whereas small coefficients for the GNN (small $\lambda_{\psi}$) cannot preserve past structural information. 

\begin{figure}[t]
\centering
\includegraphics[width=\textwidth]{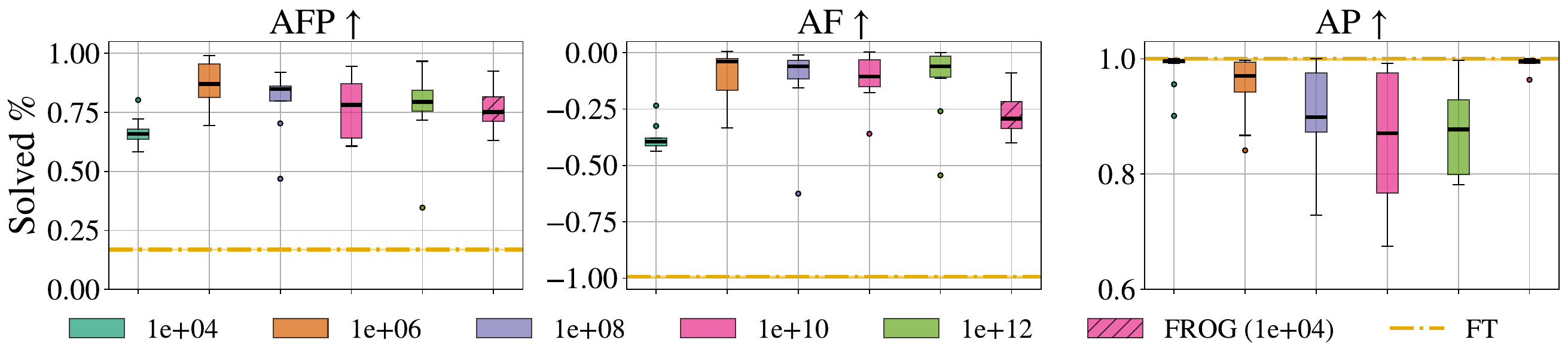}
\vspace{-0.7cm}
\caption{Boxplot showing the distribution of $10$ independent seeds in the sensitivity analysis of $\lambda_\mu$ for EWC (M) in the TS sequence. Specifically, we consider different values for the regularization coefficient $\lambda_\mu$ of the action prediction module. The regularization coefficient of the graph encoder is fixed to $\lambda_\psi =10^{10}$. When the coefficient for the action module is $\lambda_\mu =10^{10}$ (pink box), the results coincide with those reported in~\autoref{fig:full_benchmark}. We also report the results achieved by FROG, where we remark that the graph encoder is kept fixed, i.e., the coefficient $\lambda_\mu=10^4$ regularizes only the action prediction module. Boxes represent the IQR, horizontal lines indicate medians, and whiskers extend to $1.5\times\text{IQR}$.}
\label{fig:impact_reg_coeff}
\end{figure}

The results in~\autoref{fig:impact_reg_coeff} support this hypothesis: the best tradeoff is achieved with $\lambda_{\mu} = 10^6$, which is much lower than the value associated with the graph encoder ($\lambda_{\psi} = 10^{10}$). Specifically, smaller regularization coefficients for the action module increase plasticity, but may suffer from forgetting for very low values (in this case, $\lambda_{\mu} =10^4$). Conversely, strong regularization (large $\lambda_{\mu}$) causes plasticity loss, i.e., the model struggles to learn new tasks. In principle, finding the right balance between the two coefficients could be done by tuning separate values, but this may be computationally prohibitive as it requires a larger grid search. Additionally, the results achieved in this sensitivity analysis were obtained by evaluating each configuration on the entire TS sequence, which is not representative of practice. Indeed, hyperparameter tuning is usually far more restrictive: the best configuration can typically be selected only through the performance on the first few tasks~\citep{mesbahi2025position}, and we adopted this more realistic setting in the main experiments (further details in~\autoref{app:hyperparameters}). Therefore, this result further supports adopting FROG to address forgetting in CGMARL sequence: since it uses a frozen graph encoder, only $\lambda_{\mu}$ needs to be tuned, also allowing for a much lower regularization coefficient ($\lambda_{\mu} =10^4$). In this way, FROG can maintain plasticity (high AP) while achieving competitive performance (high AFP). 

\subsection{Comparison with Untrained FROG}~\label{app:freeze_from_zero}
As stated in~\autoref{sec:FROG}, FROG relies on a frozen graph backbone to mitigate forgetting arising from structural changes. To achieve this, a possible design choice consists of freezing the graph encoder \textit{before} learning the first task, i.e., after initializing the network. This is more general, as it does not adapt the GNN to a specific graph, and has proven to be effective in continual graph learning problems~\citep{donghi2025unreasonable}. Nonetheless, relying on random graph features may not provide enough expressiveness to solve MARL tasks, as agents must explore possibly large state spaces using random signals only. This may hinder learning, and for this reason, the experiments reported in~\autoref{sec:experiments} are done by freezing the graph encoder \textit{after} learning the first task. This enables learning an encoder that generates features that may also be beneficial for solving future tasks. 

\begin{figure}[t]
\centering
\includegraphics[width=\textwidth]{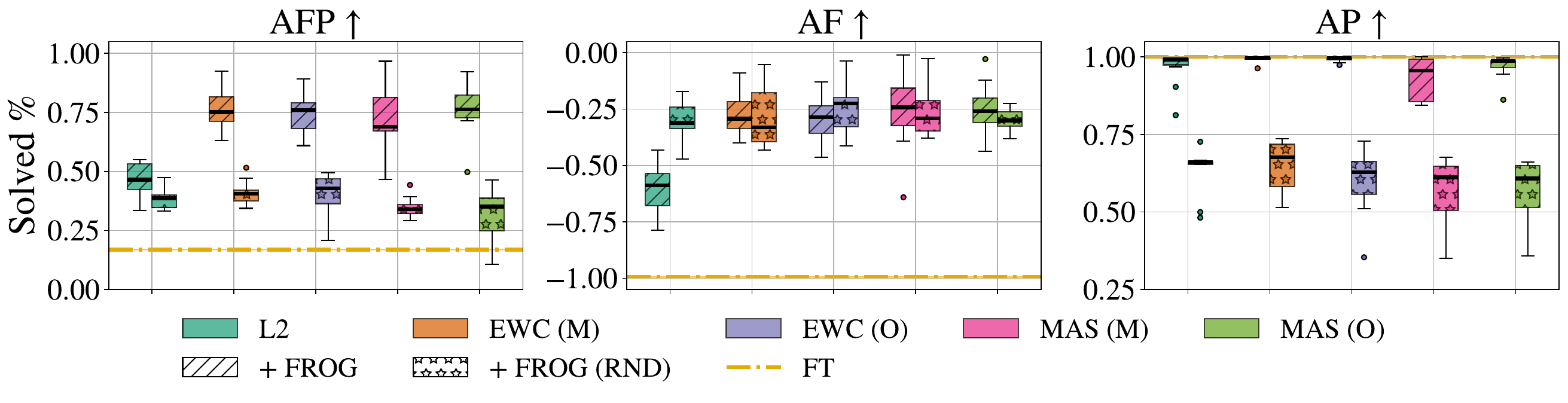}
\vspace{-0.7cm}
\caption{Boxplot showing the distribution of $10$ independent seeds for FROG and FROG (RND) combined with different CL methods in the TS scenario. Boxes represent the IQR, horizontal lines indicate medians, and whiskers extend to $1.5\times\text{IQR}$.}
\label{fig:ablation_frog}
\end{figure}

In this section, we compare FROG against the untrained variant (FROG (RND)), where the graph encoder is frozen at initialization, and the generated graph features are random. Results in~\autoref{fig:ablation_frog} show that, for all CL models, FROG (RND) struggles to learn most tasks (low AP values), thus consistently achieving worse performance (AFP) compared to FROG. This supports our preliminary hypothesis that, in GRAFO, random graph features may not be expressive enough to solve hard tasks.

\subsection{Impact of Task Identities}~\label{sec:task_id_ablation}
\begin{figure}[t]
\centering
\includegraphics[width=\textwidth]{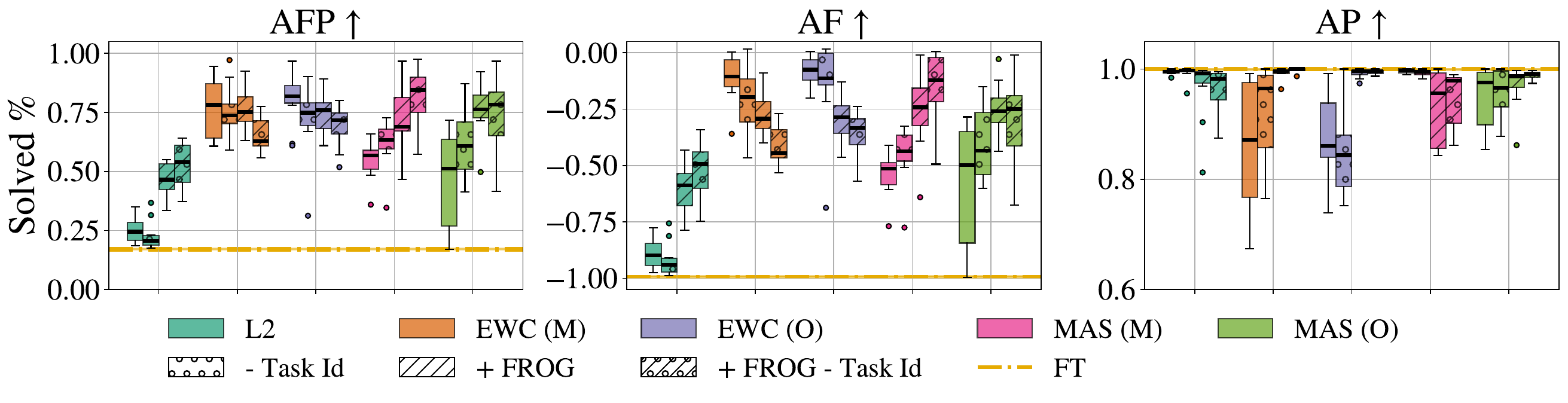}
\vspace{-0.7cm}
\caption{Boxplot showing the distribution of $10$ independent seeds for different CL methods in the TS scenario with and without conditioning on task identities. Boxes represent the IQR, horizontal lines indicate medians, and whiskers extend to $1.5\times\text{IQR}$.}
\label{fig:one_hot_id}
\end{figure}

In our experiments~(\autoref{sec:experiments}), we adopted the task-incremental learning paradigm, where we used separate task-specific heads and conditioned the observations on task identities via one-hot encoding. In~\autoref{fig:one_hot_id}, we assess the impact of task identities by comparing the results on the TS sequence with the variant that does not use the one-hot encoded identities as input. The AFP metric shows that removing task identities does not alter the overall trend when comparing standard CL methods with FROG. Within each method, consistently with~\citet{tomilin2026meal}, results remain compatible, with performance differences falling mostly within the uncertainty range. 

\section{Experimental Setting}~\label{app:exp_app}

\subsection{InforMARL}~\label{app:informarl}
InforMARL~\citep{nayak2023scalable} is a multi-agent actor-critic method based on multi-agent PPO~\citep{yu2022surprising} that leverages a GNN to exchange messages across connected agents. Specifically, at each step $t$, each agent takes an action $\va^i_t$:
\begin{equation}
    \va_t^i = \mu(\vo_t^i || \vx_{agg}^i),\quad\text{where } \vx_{agg}^i = \psi(\{\vo^i\}_{i\in\gN};\gG)
\end{equation}
where the GNN encoder $\psi$ follows the Unified Message Passing Model (UniMP;~\citep{shi2020masked}) to share messages and update the node representation:
\begin{equation}
    \vx_i' = W_1 \vx_i + \sum_{j\in\gB(i)}\alpha_{i,j} W_2\vx_j,\quad\text{where } \alpha_{i,j} = \text{softmax}\left(\frac{(W_3\vx_i)^T(W_4\vx_j + W_5\ve_{ij})}{\sqrt{c}}\right)
\end{equation}
where $W_{k=1,2,3,4,5}$ are learnable parameter matrices, $\gB(i)$ is the set of neighbors of agent $i$, and $c$ is the output dimension of the layer.
We concatenate features at each message-passing round to generate node representations at different scales.
Conversely, the critic leverages a graph-level aggregation module to estimate the value function:
\begin{equation}
    \mX_{agg} = \frac{1}{n}\sum_{i\in \gN}\vx_{agg}^i
\end{equation}
We remark that InforMARL operates on a relative frame of reference, where each agent sees other agents' positions and landmarks with respect to its position. 

\begin{figure}[t]
\centering
\includegraphics[width=\textwidth]{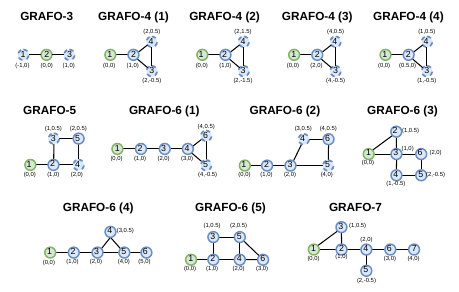}
\vspace{-0.7cm}
\caption{GRAFO tasks used to define the CGMARL sequences. The green circle is the reference node, i.e., the agent for which the target position corresponds to the reference landmark. Blue circles represent agents for which the corresponding target positions $\vg = (\vg_1, \dots,\vg_n)$ are unknown, and are reported as relative shifts (one per agent) with respect to the green node. Dashed blue nodes indicate agents in $\gK\subset\gG$ with a structurally equivalent role in the graph, i.e., that can be exchanged without altering the topology and for which a linear assignment with respect to the target positions is solved. }
\label{fig:grafo_tasks}
\end{figure}

\subsection{CGMARL Sequences}~\label{sec:task_sequences}
\paragraph{GRAFO Tasks}
In this Section, we report the GRAFO tasks and sequences used in our experiments. In particular, we report the single tasks in~\autoref{fig:grafo_tasks}: we indicate the reference agent for which the target position coincides with the known landmark as the green circle, while blue ones are agents for which the targets are unknown. The $(x,y)$ coordinates associated with each circle represent the relative shifts $\vg$ with respect to the known landmark. Moreover, dashed blue circles represent structurally equivalent nodes, i.e., agents for which the target position can be exchanged. 

\paragraph{Preliminary Experiments} In the experiments conducted in~\autoref{sec:forgetting_cgmarl}, we used GRAFO-$3$ and GRAFO-$5$ from~\autoref{fig:grafo_tasks}. For the continual learning (CONT.) experiment, we considered the following sequences:

\begin{align*}
\text{GRAFO-}(3,5)  &= [\text{GRAFO-}3;\text{GRAFO-}5]\\
\text{GRAFO-}(5,3)  &= [\text{GRAFO-}5;\text{GRAFO-}3]
\end{align*}

\paragraph{Main Experiments} For the main experiments (\autoref{sec:experiments}), we define CGMARL sequences using the graphs from~\autoref{fig:grafo_tasks}. Specifically, the sequences used in our experiments are as follows:
\begin{align*}
    \gT_{TS} &= [\text{GRAFO-}5; \text{GRAFO-}6(1); \text{GRAFO-}7; \text{GRAFO-}4; \text{GRAFO-}3; \text{GRAFO-}6(3) ]\\
    \gT_{TS\text{ (inc)}} &= [\text{GRAFO-}3; \text{GRAFO-}4; \text{GRAFO-}5; \text{GRAFO-}6(1); \text{GRAFO-}6(3);\text{GRAFO-}7  ]\\
    \gT_{TS\text{ (dec)}} &= [\text{GRAFO-}7; \text{GRAFO-}6(3); \text{GRAFO-}6(1);\text{GRAFO-}5;  \text{GRAFO-}4; \text{GRAFO-}3 ]\\
    \gT_{CS} &= [\text{GRAFO-}6(1);\text{GRAFO-}6(2);\text{GRAFO-}6(3); \text{GRAFO-}6(4);\text{GRAFO-}6(5)]\\
    \gT_{GFS} &= [\text{GRAFO-}4(1);\text{GRAFO-}4(2);\text{GRAFO-}4(3); \text{GRAFO-}4(4)]
\end{align*}
We remind that $\gT_{TS\text{ (inc)}}$ and $\gT_{TS\text{ (dec)}}$ are defined by sorting the tasks of the TS sequence $\gT_{TS}$ in incremental and decremental order with the number of agents. Moreover, the CS setting is characterized by a fixed number of nodes, which is fixed to $6$ agents for the sequence $\gT_{CS}$. Lastly, the GFS setting uses a fixed topology and varies only the graph features. For the $\gT_{GFS}$ sequence, we consider a $4$-agent graph with different target distances $\vg$, and encode the $L^2$-distances between target positions of connected nodes as edge features.

\subsection{Evaluation Metrics}~\label{app:metrics}
Given a task sequence $\gT$ with $K$ tasks, let us denote $R_{i,j}$ as the evaluation performance on task $j$ after training on task $i$. In GRAFO, this value is the percentage of solved tasks during evaluation, i.e., the percentage of episodes in which the agent team successfully managed to reach the target configuration. To evaluate the models, we use CL metrics commonly employed in the literature~\citep{wolczyk2021continual,caccia2022new,sokar2025continual}.
The metrics of interest are defined as follows.
\paragraph{Average Final Performance} The Average Final Performance (AFP) measures the model's performance across all tasks after learning the last one:
    \begin{equation}
        AFP = \frac{1}{K}\sum_{j=1}^K R_{K,j} 
     \end{equation}
\paragraph{Average Forgetting} The Average Forgetting (AF) measures how performance on past tasks deteriorated after learning the final task (backward transfer). It is defined as:
\begin{equation}
       AF = \frac{1}{K-1}\sum_{j=1}^{K-1}(R_{K,j} - R_{j,j} )
\end{equation}

\paragraph{Average Plasticity} The Average Plasticity (AP) measures the performance on a task immediately after learning it. It is defined as:
\begin{equation}
       AP = \frac{1}{K}\sum_{j=1}^{K}R_{j,j}
\end{equation}

\subsection{Implementation Details}~\label{app:implementation}
\subsubsection{Software and Hardware}
We implemented the GRAFO benchmark on top of the \textit{Jax} version~\citep{JaxInforMARL}~\footnote{https://github.com/jselvaraaj/JaxInforMARL} of InforMARL~\citep{nayak2023scalable}. To implement the continual learning methods, we adapted parts of the code from the repository~\footnote{https://github.com/TTomilin/MEAL} of the MEAL benchmark~\citep{tomilin2026meal}.  Experiments for this research were conducted on the HPC cluster at Università della Svizzera italiana and the Intelligent Autonomous Systems (IAS) cluster at TU Darmstadt.

\subsubsection{Hyperparameters}~\label{app:hyperparameters}
We report the hyperparameters used in our experiments in~\autoref{tab:hyperparameters}. To select them, we first tuned the parameters of the GNN-based policy (InforMARL) on a single task. These are kept fixed for all the CL methods, both with and without FROG.~\autoref{tab:TS_single_task_resolution} in~\autoref{app:single_task_performance} show that the chosen configuration can solve each task individually. Then, we followed $k$-percent tuning~\citep{mesbahi2025position} to choose the parameters for each different CL method. Specifically, for each configuration, we ran $3$ seeds on the first $2$ tasks of the TS sequence, and selected the hyperparameters with the best average final performance. 

\def\thickhline{\noalign{\hrule height1.3pt}}

\begin{table}[h!]
\vspace{-0.3cm}
\caption{Hyperparameters used in our experiments. We note that the parameters specific to the online variants of EWC and MAS (with and without FROG), i.e., online decay, importance episodes, importance steps, and normalization, are common for all methods.}
\vspace{0.2cm}
\label{tab:hyperparameters}
\renewcommand{\arraystretch}{1.1}
\centering
{
\begin{tabular}{c|c|c|c}
\thickhline
Context & Parameter & Value   \\
\thickhline
\parbox[t]{2mm}{\multirow{12}{*}{\rotatebox[origin=c]{90}{Architecture (InforMARL)}}} & Activation  & $ReLU$\\
& Num. of hidden layers (action module) & $2$\\
& Hidden feature dim (action module) & $128$ \\
& Recurrent unit (action module) & False \\
& Num. of hidden layers (critic) & $1$\\
& Hidden feature dim (critic) & $128$ \\
& Recurrent unit (critic) & True \\
& Num. of hidden layers (GNN) & $2$ \\
& Hidden feature dim (GNN) & $128$ \\
& Num. of graph attention layers (GNN) & $4$ \\
& Num. of attention heads (GNN) & $3$\\
& Graph attention key dim (GNN) & $16$\\
\hline
\parbox[t]{2mm}{\multirow{9}{*}{\rotatebox[origin=c]{90}{Optimization (PPO)}}}& Clip $\epsilon$ & $0.2$\\
&Learning rate &  $5\cdot10^{-4}$\\
&Update epochs &  $4$\\
&Discount factor &  $0.99$\\
&GAE $\lambda$ &  $0.95$\\
&Entropy coeff. &  $0.001$\\
&Value coeff. &  $0.01$\\
&Rollout Length &  $128$\\ 
&Minibatches per update &  $32$\\ 
\hline
\parbox[t]{2mm}{\multirow{5}{*}{\rotatebox[origin=c]{90}{\small{Env. (GRAFO)}}}}&Env. steps per task &  $5\cdot 10^6$\\
&Episode length &  $50$\\
&Parallel environments &  $32$\\
&Evaluation episodes & $64$\\
&Enable collisions & False \\ 
\hline
\parbox[t]{2mm}{\multirow{11}{*}{\rotatebox[origin=c]{90}{{Continual Learning}}}}&Reg. coeff. (L2 / L2 + FROG) & $10^3$ / $10^2$\\ 
&Reg. coeff. (EWC (M) / EWC (M) + FROG) & $10^{10}$ / $10^4$ \\ 
&Reg. coeff. (EWC (O) / EWC (O) + FROG) & $10^{11}$ / $10^5$ \\  
&Reg. coeff. (MAS (M) / MAS (M) + FROG) & $10^{7}$ / $10^6$ \\ 
&Reg. coeff. (MAS (O) / MAS (O) + FROG) & $10^{9}$ / $10^4$ \\ 
&Regularize critic &  False\\ 
&Regularize head &  False\\ 
&Online decay&  $0.9$\\
&Importance episodes &   $64$\\ 
&Importance steps &  $50$\\ 
&Normalize importance &  False\\ 
\hline
\end{tabular}
}
\end{table}

\end{document}